%% file: main.tex
\documentclass[sigconf]{acmart}
\usepackage{shortcuts}
\usepackage{bm}
\usepackage{algorithm,algorithmic}
\usepackage{makecell}
\usepackage{float}
\usepackage{wasysym}
\usepackage{tcolorbox}
\usepackage{array}
\usepackage{multirow}

\usepackage[skip=2pt]{caption}
\usepackage{graphicx}
\usepackage{booktabs}
\usepackage[misc]{ifsym}

\usepackage{resizegather}
\usepackage{threeparttable}
\usepackage{tikz}
\usepackage{mathtools}
\usepackage{subcaption}
\usepackage{ragged2e}

\AtBeginDocument{%
  }

\copyrightyear{2024}
\acmYear{2024}
\setcopyright{acmlicensed}
\acmConference[CCS '24]{Proceedings of the 2024 ACM SIGSAC Conference on Computer and Communications Security}{October 14--18, 2024}{Salt Lake City, UT, USA}
\acmBooktitle{Proceedings of the 2024 ACM SIGSAC Conference on Computer and Communications Security (CCS '24), October 14--18, 2024, Salt Lake City, UT, USA}
\acmDOI{10.1145/3658644.3690292}
\acmISBN{979-8-4007-0636-3/24/10}

\begin{document}

\title{Uncovering Gradient Inversion Risks in Practical Language Model Training}


\author{Xinguo Feng}
\affiliation{%
  \institution{The University of Queensland}
  \streetaddress{St Lucia}
  \city{Brisbane}
  \country{Australia}
  }
\email{s.feng@uq.edu.au}

\author{Zhongkui Ma}
\affiliation{%
  \institution{The University of Queensland}
  \streetaddress{St Lucia}
  \city{Brisbane}
  \country{Australia}}
\email{zhongkui.ma@uq.edu.au}

\author{Zihan Wang}
\affiliation{%
  \institution{The University of Queensland}
  \streetaddress{St Lucia}
  \city{Brisbane}
  \country{Australia}}
\email{zihan.wang@uq.edu.au}

\author{Eu Joe Chegne}
\affiliation{%
  \institution{The University of Queensland}
  \streetaddress{St Lucia}
  \city{Brisbane}
  \country{Australia}}
\email{e.chegne@uqconnect.edu.au}

\author{Mengyao Ma}
\affiliation{%
  \institution{The University of Queensland}
  \streetaddress{St Lucia}
  \city{Brisbane}
  \country{Australia}}
\email{mengyao.ma@uq.edu.au}


\author{Alsharif Abuadbba}
\affiliation{%
  \institution{CSIRO's Data61}
  \streetaddress{Marsfield}
  \city{Sydney}
  \country{Australia}}
\email{sharif.abuadbba@data61.csiro.au}

\author{Guangdong Bai}
\authornote{Corresponding author}
\affiliation{%
  \institution{The University of Queensland}
  \streetaddress{St Lucia}
  \city{Brisbane}
  \country{Australia}}
\email{g.bai@uq.edu.au}

\renewcommand{\shortauthors}{Xinguo Feng et al.}

\input{sections/abstract_cr}




\maketitle

\input{sections/1_intro_cr}
\input{sections/2_related_work_cr}
\input{sections/3_preliminaries_cr}
\input{sections/4_approach_cr}
\input{sections/5_experiment_cr}
\input{sections/6_conclusion_cr}

\section*{Acknowledgments}
We thank our anonymous shepherd and reviewers for their constructive comments.
This work is partially supported by Australian Research Council Discovery Projects~(DP230101196, DP240103068).

\bibliographystyle{ACM-Reference-Format}
\bibliography{references}
\appendix
\input{sections/appendix_A}

\input{sections/appendix_B}










\end{document}

%% file: sections/abstract_cr.tex
\begin{abstract}

The {gradient inversion} attack has been demonstrated as a significant privacy threat to federated learning~(FL), particularly in continuous domains such as vision models. 
In contrast, it is often considered less effective or highly dependent on impractical training settings when applied to language models, due to the challenges posed by the \emph{discrete} nature of tokens in text data. 
As a result, its potential privacy threats remain largely underestimated, despite FL being an emerging training method for language models. 
In this work, we propose a domain-specific gradient inversion attack named \codename~(\textbf{\underline{gra}}dient inversion with \emph{hy\textbf{\underline{b}}rid} optimization). 
\codename features two alternating optimization processes to address the challenges caused by practical training settings, including a simultaneous optimization on dropout masks between layers for improved token recovery and a discrete optimization for effective token sequencing.
\codename can recover a significant portion~(up to 92.9\% recovery rate) of the private training data, outperforming the attack strategy of utilizing discrete optimization with an auxiliary model
by notable improvements of up to 28.9\% recovery rate in benchmark settings and 48.5\% recovery rate in practical settings.
\codename provides a valuable step forward in understanding this privacy threat in the emerging FL training mode of language models. 

\end{abstract}

%% file: sections/1_intro_cr.tex
\section{Introduction}
\label{sec:introduction}

The advent of neural network-based language models represents a significant breakthrough in the field of natural language processing~(NLP), enabling remarkable performance in various tasks such as named entity recognition~\cite{ner}, sentiment analysis~\cite{sentiment}, text summarization~\cite{textsum} and text generation~\cite{gpt4, instructGPT}. 
Nevertheless, the requirement of ever-larger models and datasets poses a challenge to model training\footnote{In practice, training a language model typically refers to fine-tuning a published pre-trained language model with a local training dataset for a downstream task. Without loss of generality, we use ``training'' to refer to this practice.}---the need for large text corpus and data security.
Federated learning~(FL) \cite{fl1} steps in this context as a pivotal solution. By facilitating collaborative training over distributed data while ensuring the ownership of individual training data, it enables multiple clients to collaborate to train a model without sharing their raw data with other entities.
As a result, the use of FL in language model training has attracted much attention from both academia, \textit{e.g.,} FedFSL~\cite{cai2023federated} and FedBERT~\cite{tian2022fedbert}, and industry, \textit{e.g.,} FedML~\cite{FedML}.

Despite the training data being confined locally, recent studies~\cite{dlg,idlg} have revealed that an adversary can reconstruct the private training data of a client via an attack known as \textit{gradient inversion}. 
This attack recovers training data by randomly initializing dummy data and updating it with an optimization process through gradient descent to produce similar model parameter gradients, hence approximating the original training data.
The power of gradient inversion is largely due to the efficacy of the optimization techniques, and therefore it shows a high efficacy for the \emph{continuous} data, such as adjusting the pixel values of the surrogate images to mirror the original images in vision models~\cite{dlg,idlg,april,ig, gip,stg}. 
However, this strength wanes when it comes to language models, a domain characterized by the \emph{discrete} nature of tokens. 

Several efforts have been dedicated to adapting gradient inversion for language models. 
One line of research~\cite{dlg, tag} explores applying attack strategies originally designed for continuous domains onto token embedding, treating embedding as the ``continuous'' representation of tokens. 
This retains the capability of gradient inversion in recovering tokens, but falls short in restoring token order due to the dominance of token embedding over positional embedding. 
Attack strategies based on discrete optimization~\cite{lamp} seem promising in coping with the token discreteness. 
However, due to the challenges posed by dropout-induced randomness and long-sequence tokens reordering, they have to strongly rely on controlled experimental settings such as deactivating dropout and providing exact sequence lengths.
\emph{The applicability of gradient inversion in practical language model training scenarios} remains largely open. 
Consequently, the privacy threats posed by gradient inversion against language models have been underestimated. 

\paragraph{Our work}
We introduce a \textbf{\underline{gra}}dient inversion with a hy\textbf{\underline{b}}rid optimization called \codename, an attack specifically for language model training in practical FL scenarios.
\codename's hybrid optimization combines the continuous optimization, which is highly sensitive to initialization~\cite{film}, with the discrete optimization, which requires the correct recovered tokens to take effect. 
It schedules them in an iterative paradigm to complement each other, where the output of one serves as an improved initialization for the other in the new iteration. This approach helps avoid local optima and achieves better final recovery.
During each iteration, \codename considers the dropout masks between layers and uses gradient descent with dropout mask learning as the continuous optimization to offset the dropout-induced noise, mainly for effective \emph{token recovery}. 
It employs beam search on the recovered tokens as the discrete optimization for systematic \emph{token sequencing}. 
This involves the padding token to dynamically adjust the valid sequence lengths, eliminating the need for prior knowledge of the exact sequence lengths in the batch.

During the continuous optimization, \codename employs a dropout mask learning technique to counteract the negative effects caused by dropout~\cite{scheliga2023dropout}. 
Dropout is a typical regularization technique for overfitting prevention in deep learning. 
It also introduces noise to the gradient by randomly omitting a subset of neurons in every training round. 
This poses a major challenge to the traditional continuous gradient inversion, as missing gradients and potential perturbations are caused. 
\codename's dropout mask learning technique initializes a dummy dropout mask and updates it alongside the dummy data during the optimization process. This measure effectively offsets the dropout-induced noise, which aids the gradient matching and substantially results in an improved token recovery in the presence of dropout.

In the discrete optimization, \codename focuses on reordering the tokens recovered from the continuous optimization.   
The continuous gradient inversion falls short in restoring token order due to the dominance of token embedding over positional embedding~\cite{dlg,tag,lamp,film}. 
This limitation becomes particularly pronounced when no prior knowledge of the exact sequence lengths is provided to the attacker in a practical FL scenario, which introduces numerous potential orderings.
To counter this limitation, \codename employs beam search for reordering.
Unlike the existing discrete strategy~\cite{lamp} that utilizes an external auxiliary model to guide the search, beam search systematically explores the search space by iterating through each position in the sequence and utilizing the interconnection between adjacent tokens, resulting in the capability of recovering significantly longer sequences. 
Particularly, our approach involves the padding token in the reordering to dynamically adjust the valid sequence lengths, which requires no prior knowledge of the exact sequence lengths in the batch. 

To evaluate the performance of \codename, we conduct both comparative and ablation studies.
To evaluate its attack efficacy,
we first apply it to three widely-used datasets and compare it with three baseline approaches using various optimization strategies~\cite{dlg, tag, lamp} in benchmark settings.
The results demonstrate the remarkable recovery ability of \codename. 
It recovers a large portion~(up to 92.9\% recovery rate) of the training data, and surpasses the best baseline approach~\cite{lamp} that is based on the discrete optimization with an auxiliary model by notable improvements of up to 28.9\% recovery rate. 
Subsequently, to highlight the practicality of \codename, we introduce the practical settings, including frozen embedding layers and activated dropout. 
In these settings, \codename remains highly effective, outperforming the best baseline by a remarkable improvement of up to 48.5\% recovery rate. 
This shows its strong practicality. 
In ablation studies, we evaluate \codename with various training settings, including various model sizes/types~(all sizes of BERT~\cite{bert} and RoBERTa~\cite{roberta} models), various dropout rates~\cite{dropout}, more complex task (multi-class classification), large batch sizes (up to 128), and relaxed assumptions. It remains a highly effective and consistent attack performance.
Furthermore, we evaluate it against proactive defenses, including gradient noise~\cite{dlg, wei2020framework} and gradient pruning~\cite{dlg}. The results reveal its resilience against existing defenses.

\paragraph{Contributions}
Our contributions are summarized as follows.
\begin{itemize}
    \item \textbf{A novel gradient inversion in the domain of language models.}
    We introduce \codename, a novel gradient inversion with hybrid optimization to address domain-specific challenges. 
     It adapts the dropout-mask learning technique to language models for a dropout-aware continuous optimization. It then employs the beam search strategy that involves padding tokens as a systematic discrete optimization, which results in effective data recovery.
    \codename represents a step forward in understanding the privacy threats posed by the gradient inversion against the emerging FL training mode of language models. 
    
    \item \textbf{Effective attack performance}.
    \codename demonstrates a high recovery rate of private training data in both benchmark and practical settings.
    It remains highly effective across various training settings, such as model sizes/types and dropout rates.
    These findings underscore the threat posed by gradient inversion against language models, refreshing the current understanding of this attack in an emerging domain. 
        
    \item \textbf{Strong practicality and resilience.} 
    To the best of our knowledge, \codename is the first \emph{practical} gradient inversion against language models.
    It formulates key factors that cause challenges for inversion attacks, such as activated dropout, frozen embeddings, large batch sizes, and unknown sequence lengths.
    \codename achieves high attack efficacy in practical training settings and defense strategies, demonstrating its strong practicality. 
    This establishes a new baseline for gradient inversion attacks for language models, providing a valuable step forward in understanding this previously underestimated threat.
\end{itemize}

\paragraph{Notations} The notations used throughout this work are introduced. Lower-case letters, such as $b$, $d$, $i$, $j$, $l$, $m$, $n$, $p$, $s$, $t$, $\alpha$, and $\lambda$, are employed to denote scalar variables or constants.
The bold lower-case letters, such as $\bm{l}$, $\bm{x}$, $\bm{y}$, and $\bm{\theta}$, represent vectors or sequences, and bold capitalized letters, such as $\bm{X}$, $\bm{B}$, $\bm{E}$, and $\bm{\Psi}$, stand for matrices or tensors. 
We use upper-case calligraphic letters, such as $\mathcal{L}$, $\mathcal{D}$, to denote functions. 
We use $\nabla$ to denote the gradient operator.
For conciseness, the set of consecutive integers from $m$ to $n$ is referred to as $m..n$.

\paragraph{Availability} Our code is publicly available at: \url{https://github.com/UQ-Trust-Lab/GRAB/}.

%% file: sections/2_related_work_cr.tex
\section{Related Work}
\label{sec:related_work}

This section delves into existing attempts of gradient inversion attacks on language models, followed by a discussion on typical training settings and assumptions in existing attacks, and existing defense strategies.

\subsection{Existing Attacks}
\label{sec:related_work:gia}

Deep learning models trained in FL context are known to be vulnerable to privacy attacks in the FL process. Besides attacks such as membership inference ~\cite{ma2023loden, shokri2017membership, mireshghallah2022quantifying} that aim to infer if a data sample is part of the training data, 
previous studies show that gradients shared by the clients in FL can be used to recover the private training data with gradient inversion for vision models~\cite{dlg,idlg,ig,april,stg,gip}.
These attacks achieve a high recovery rate by utilizing the recovered labels, image priors, and batch normalization statistics.
Gradient inversion primarily focuses on continuous data. While image data is continuous (pixel values) and text data is discrete (individual tokens), text data can be mapped to embedding vectors through the embedding layer of language models, such that it can be considered continuous in the embedding space.
Recently, more studies show that it is feasible to apply similar methods to recover text data of language models~\cite{dlg, tag,film,lamp,decep}. 

\paragraph{Solely utilizing continuous optimization}
Initially, gradient inversion on language models relies merely on the same techniques for image data, which is highly sensitive to initialization~\cite{film}.
Zhu \textit{et al.}~\cite{dlg} propose DLG and firstly show that the same attack for vision models can be directly applied to a BERT model~\cite{bert} and recover part of its training data.
Deng \textit{et al.}~\cite{tag} propose TAG, an improved version of the original attack, which adds the $L_1$ norm of the gradient distance as a regularization term to improve the token recovery. 

\paragraph{Combining with discrete optimization}
Despite the partial success of directly utilizing continuous optimization, this approach falls short in restoring token order
due to the dominance of token embedding over positional embedding~\cite{dlg, tag, lamp, film}.
The employment of discrete techniques enhances the effectiveness of the inversion process by reordering the recovered tokens to further minimize the gradient distance. 
Inspired by the black-box data extraction attack proposed by Carlini \textit{et al.}~\cite{carlini2021extracting}, Gupta \textit{et al.}~\cite{film} propose FILM, an attack that utilizes the memorization ability of a GPT-2 model~\cite{gpt2} as a probability distribution to reconstruct the private training text sequences with the highest overall probability.
This technique is primarily for Causal Language Models (CLMs).
Most related to our work for Masked Language Models (MLMs), Balunovic \textit{et al.}~\cite{lamp} propose LAMP, an attack that alternates gradient descent as continuous optimization and random search as discrete optimization with the aid from an external auxiliary language model to recover private training text data of a BERT~\cite{bert} model.
However, its success deeply relies on settings such as deactivated dropout and prior knowledge of exact sequence lengths in the batch.
We emphasize that CLMs are easier to attack, as the auto-regressive nature of the model can be utilized as a probability distribution on the next possible tokens, while the MLMs do not have such characteristics.

\paragraph{Active attack} In addition to the passive attacks mentioned above, Fowl \textit{et al.}~\cite{decep} propose an attack called DECEPTICONS that deploys malicious parameters vector to recover high-fidelity private text from a GPT-2 model~\cite{gpt2}. However, this attack is not evaluated against any active detection defense and its stealthiness is unclear.

\subsection{Training Settings and Assumptions in Existing Attacks}
Most existing attacks are based on certain training settings and assumptions. 
Table~\ref{table:assumptions} shows the training settings, assumptions, method combinations of \codename and the related approaches that target MLMs.

\input{tables/assumptions}

\subsubsection{Training Settings in Existing Attacks}\label{section:settings}

Common training settings in existing attacks include trainable embedding layers and deactivated dropout.

\paragraph{Trainable embedding layers}
DECEPTICONS~\cite{decep} and FILM~\cite{film} follow the observation from Melis \textit{et al.}~\cite{melis}, which shows that the non-zero rows of token embedding gradients can actually reveal the tokens that are within the training batch.
Lu \textit{et al.}~\cite{april} propose APRIL and demonstrate that with a batch size of 1, the positional embedding layer gradients can be utilized to reveal the full sequence. 

This leakage from embedding layers has led to the idea of freezing learnable embedding, resulting in no gradients being generated for the parameters in these layers to defend the attacks~\cite{film,april}.
This lightweight defense is shown to be effective against attacks that directly utilize embedding layer gradients, without sacrificing the utility of the model.
It is recommended to be applied before the training starts~\cite{film}.
However, in current attacks, the embedding layers are still trainable, which might aid the data recovery even if the gradients of these layers are not directly utilized for direct token leakage.

\paragraph{Deactivated dropout}
Dropout~\cite{dropout} is a commonly employed countermeasure against overfitting in deep learning.
Fundamentally, dropout employs a dropout mask, which controls the participation of neurons during training, subsequently influencing the computed gradients by introducing randomness and perturbing the gradient patterns.
This mechanism is the most influential factor in the effectiveness of continuous optimization in a practical FL scenario~\cite{scheliga2023dropout}.
Upon further inspection of the implementation of existing attacks~\cite{dlg, tag, lamp}, it is common that dropout is deactivated during training. The attacker can facilitate gradient descent for token recovery without considering the dropout-induced noise, which is usually not the case in a practical FL scenario. 

Recent studies~\cite{wei2020framework} suggest that dropout can be used as a defense mechanism to mitigate gradient inversion attacks. 
Utilizing active dropout masks introduces a challenge for the attacker attempting to invert embeddings through gradients, primarily because the attacker lacks insight into the specific dropout mask realization of the victim client.
However, in the study by Scheliga \textit{et al.}~\cite{scheliga2023dropout}, a technique is introduced that incorporates dropout mask learning within the gradient inversion.
This method, when applied to a Vision Transformer (ViT) model~\cite{vit}, demonstrates substantial attack effectiveness, even when faced with elevated dropout rates, as the learned dropout mask further aids the optimization to counteract missing gradients and the dropout-induced noise.

\subsubsection{Assumptions in Existing Attacks}
\label{section:related_work:assumptions}

Most existing attacks are based on certain assumptions on the knowledge of the adversary~\cite{dlg,tag,lamp}.
For text data, the usual assumptions are the labels and the exact sequence lengths in the batch are known to the attacker~\cite{lamp}.
For the labels, existing attacks argue that they can apply the method introduced by Zhao \textit{et al.}~\cite{idlg} to identify the labels present in the batch, followed by enumerating all combinations. However, this approach only identifies the presence of specific labels but fails to ascertain the quantity of each label or the precise label associated with each sequence in the batch.
For the sequence lengths, existing attacks also justify that they can enumerate all length combinations in the batch. Both assumptions become impractical as the batch size grows, as the number of possible combinations is enormous.

\subsection{Defenses against Gradient Inversion Attacks}
\label{sec:related_work:defense}

Gradient inversion relies on minimizing the gradient distance between the dummy gradients and the shared gradients to approximate the original training data. To mitigate the effectiveness of such attacks, existing defenses primarily focus on perturbing gradient patterns or injecting noise into the gradients. Specifically, two main approaches are employed.

\paragraph{Freezing embedding layers and activating dropout}
As discussed above in Section~\ref{section:settings}, freezing embedding layers and activating dropout can serve as lightweight defense mechanisms.
Freezing embedding layers can prevent direct gradient information leakage of tokens involved in the batch, while activated dropout can perturb the gradients by omitting a subset of neurons.

\paragraph{Pruning or perturbing gradients}
Pruning or perturbing gradients is another defense strategy against gradient inversion attacks by slightly modifying the original gradients.
Zhu \textit{et al.}~\cite{dlg} propose a defense that prunes the gradients of small magnitudes to perturb the gradient patterns.
Additionally, another approach suggests directly injecting Gaussian or Laplacian noise into the gradient to further perturb them~\cite{dlg, wei2020framework}.

%% file: tables/assumptions.tex
\begin{table}[t]
\centering
\newcommand{\twocol}[1]{\multicolumn{2}{c}{#1}} 
\renewcommand{\arraystretch}{1.1}
\caption{Comparisons of settings, assumptions, and methods of different approaches.}
\label{table:assumptions}

\begingroup 
\setlength{\tabcolsep}{1.7pt}
\begin{threeparttable}
\scriptsize
\begin{tabular}{lcccccc} 
\toprule
& \twocol{Settings} & \twocol{Assumptions} & \twocol{Methods} \\
\cmidrule(l{5pt}r{5pt}){2-3}
\cmidrule(l{5pt}r{5pt}){4-5}
\cmidrule(l{5pt}r{5pt}){6-7}
& \makecell[c]{Trainable\\embeddings} & 
\makecell[c]{Deactivated\\dropout} & 
\makecell[c]{Known\\Seq. lengths} & 
\makecell[c]{Known\\labels} &
\makecell[c]{Continuous\\optimization} &\makecell[c]{Discrete\\optimization}\\
\midrule
DLG~\cite{dlg} & \CIRCLE & \CIRCLE & \CIRCLE & \CIRCLE & \checkmark & \\
TAG~\cite{tag} & \CIRCLE & \CIRCLE & \CIRCLE & \CIRCLE & \checkmark & \\
LAMP~\cite{lamp} & \CIRCLE & \CIRCLE & \CIRCLE & \CIRCLE & \checkmark & \checkmark \\
\codename & \Circle & \Circle & \LEFTcircle & \LEFTcircle & \checkmark & \checkmark \\
\bottomrule
\end{tabular}
\begin{tablenotes}
    \scriptsize
    \item[*] \CIRCLE~(or \Circle) refers to the presence~(or absence) of a setting or assumption.
    \LEFTcircle~ presents the assumption is optional.
    \checkmark presents the involvement of a method.
    Seq. is the abbreviation for the word ``Sequence''.
\end{tablenotes}
\end{threeparttable}
\endgroup
\end{table}





%% file: sections/3_preliminaries_cr.tex
\section{Preliminaries}
\label{sec:preliminaries}

\begin{figure*}[!t]
    \centering
    
    \includegraphics[width=0.9\linewidth]{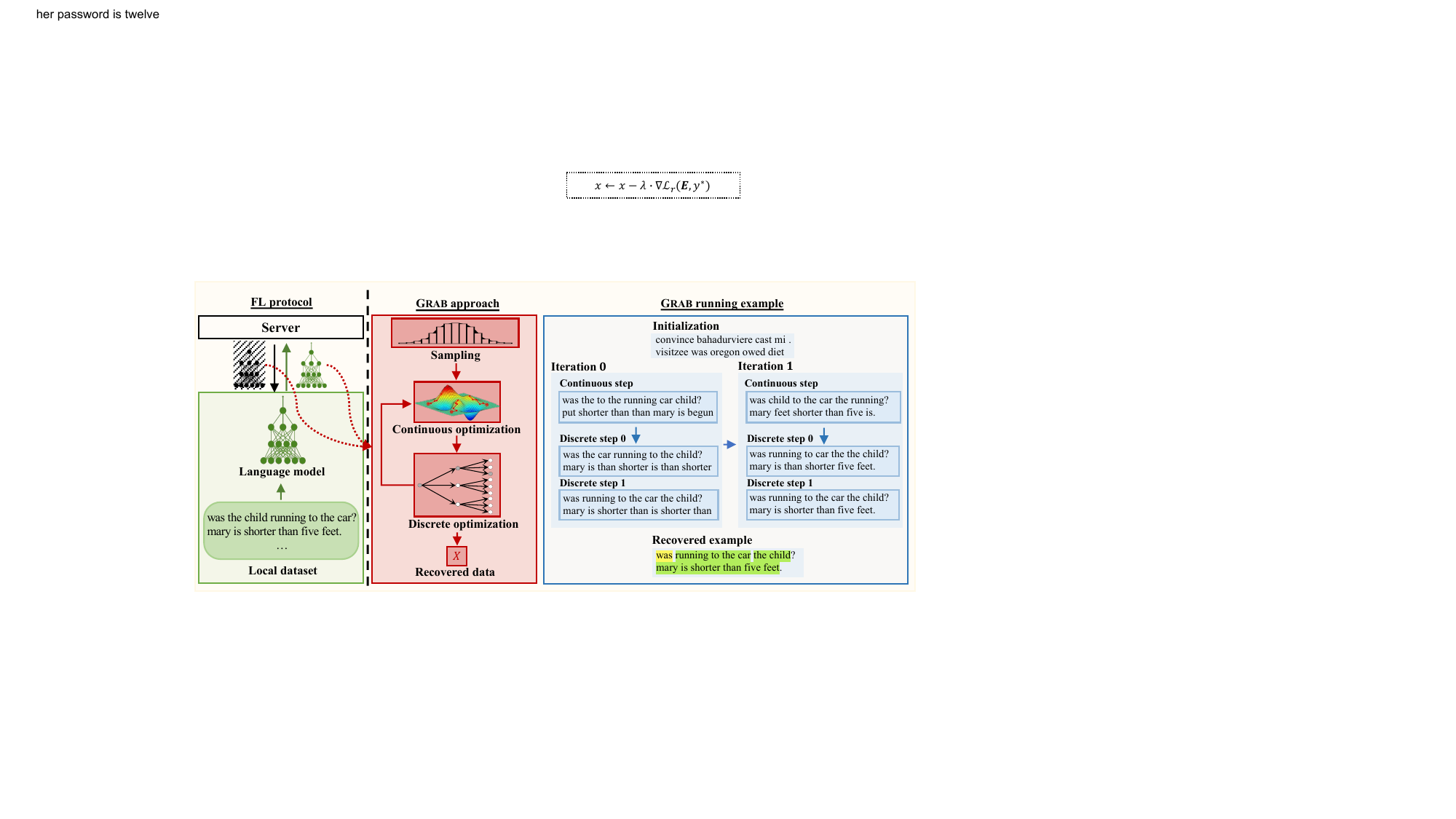}
    \caption{Overview of \codename.}
    \label{fig:overview}
\end{figure*}

In this section, we introduce the necessary preliminary knowledge to facilitate a comprehensive understanding of \codename.

\subsection{Language Modeling}

Language modeling has become a major focus in NLP due to the development of neural networks. 
This paper focuses on the MLM, which is a key framework in language modeling.  
BERT~\cite{bert}, a typical MLM, is a transformer encoder~\cite{transformer} that contains millions of parameters pre-trained on a large corpus of text. It is able to generate contextual encoding representation of text after pre-training.
Generally, a custom layer is added to the encoder and is fine-tuned for a downstream task, such as sentiment analysis~\cite{sentiment}.

In this work, we investigate a BERT model augmented with a fully connected layer for a sequence classification task.
The input is a data batch $\bm{B} = (\bm{X}, \bm{y})$, where $\bm{X} \in \mathbb{R}^{b \times n}$ is a batch of $b$ sequences with $n$ tokens and $\bm{y} \in \mathbb{Z}^b$ is a batch of the correct labels for the sequences.
Initially, $\bm{X}$ is passed through the token embedding layer to obtain the embedding $\bm{E} \in \mathbb{R}^{b \times n \times d}$, where $d$ is the embedding dimension.
Subsequently, $\bm{E}$ is fed through the remaining layers to generate the logit outputs.
The loss function $\mathcal{L}$ computes the loss value based on the softmax outputs of the logits and the labels.
Finally, the gradients $\nabla_{\bm{\theta}^t} \mathcal{L}(\bm{B})$ of parameters $\bm{\theta}^t$ at iteration $t$ are computed by backpropagation and are used to compute the updated parameters $\bm{\theta}^{t+1}$.
By repeating this process, the loss is minimized and the accuracy of the model is improved.

\subsection{Federated Learning}

Federated learning~\cite{fl1} is a machine learning framework that allows multiple clients to collaborate in training a model without having to share their individual private training data, but only the training information such as gradients, with the coordination of a server.

In this work, we consider the FedSGD~\cite{fl1} setting.
In the $t$-th training iteration, the $i$-th client computes the gradients $\nabla_{\bm{\theta}^t}\mathcal{L}(\bm{B}_i^*)$ of parameters $\bm{\theta}^t$ by the loss function $\mathcal{L}$ and local data batch $\bm{B}_i^*$.
Then, all clients send their gradients to the server.
The server computes the updated parameters $\bm{\theta}^{t+1}$ by aggregating all clients' gradients and subtracting it from the current parameters $\bm{\theta}^t$, 
\begin{equation*}\small
    \bm{\theta}^{t+1} = \bm{\theta}^t - \frac{\alpha}{m}\sum_{i=1}^m\nabla_{\bm{\theta}^t} \mathcal{L}(\bm{B}_i^*),
\end{equation*}
where $m$ is the number of clients and $\alpha$ is the learning rate.
The server then sends the updated parameters to each client for updating their local models.

\subsection{Gradient Inversion Attacks}
\label{gia}

The gradient inversion attack~\cite{dlg,idlg,tag,ig,decep,film,lamp} is a technique used to recover the training data of a client from the transmitted gradient information in FL.
It is assumed that the adversary is able to eavesdrop on the communication between the server and the victim client, and has full knowledge of the model architecture and parameters.
To recover the input data batch $\bm{B}^*_i$, an alternative data batch $\bm{B}_{i}$ is randomly initialized and optimized by solving the following optimization problem, 
\begin{equation*} \small
    \underset{\bm{B}_{i}}{\argmin} ~\mathcal{D}(\nabla_{\bm{\theta}^t} \mathcal{L}(\bm{B}_{i}),~ \nabla_{\bm{\theta}^t} \mathcal{L}(\bm{B}^*_i)), 
\end{equation*} 
where $\mathcal{D}$ is a distance measure function.
For text data, $\bm{B}_{i}$ represents the embedding of the data batch. It is further mapped back to discrete tokens by comparing it with the token embeddings of the language model.

\subsection{Droupout Mask Learning}
To counter the perturbation introduced by dropout, we adopt the dropout mask learning technique proposed by Scheliga \textit{et al.}~\cite{scheliga2023dropout}. 
Dropout mask learning aims to learn the dropout mask realization of the victim client's model in that training round to reduce the perturbation effect. Specifically, the attacker initializes the dropout masks between the layers and optimizes them simultaneously in the continuous optimization process. The optimization can be further expressed as
\begin{equation*} \small
    \underset{\bm{B}_{i}, \bm{\Psi}_{p}}{\argmin} ~\mathcal{D}(\nabla_{\bm{\theta}^t} \mathcal{L}_{\bm{\Psi}_p}(\bm{B}_{i}),~ \nabla_{\bm{\theta}^t} \mathcal{L}_{\bm{\Psi}^*_p}(\bm{B}^*_i)), 
\end{equation*}
where $\bm{\Psi}_{p}$ is the dropout mask realization being optimized.

\subsection{Beam Search}
To reorder the tokens recovered by continuous optimization, we adopt beam search as a discrete optimization technique. Beam search is a greedy-like discrete optimization, where greedy search only focuses on the current best solution, beam search keeps track of the \emph{top-k} solutions in a breadth-first-search manner along the process and selects the best solution at the end~\cite{beamsearch2}.

%% file: sections/4_approach_cr.tex
\section{Approach}
\label{sec:approach}

In this section, we give a detailed description of our proposed gradient inversion with hybrid optimization, which iteratively alternates continuous and discrete optimization to recover the private training data.
We first discuss the threat model in Section~\ref{sec:approach:threat_model}, and provide an overview of our approach in Section~\ref{sec:approach:overview}. 
The continuous and discrete optimizations are discussed in Section~\ref{sec:approach:continuous_opt} and Section~\ref{sec:approach:discrete_opt}.

\subsection{Threat Model}
\label{sec:approach:threat_model}

We consider an honest-but-curious adversary that controls the server or eavesdrops on the communication between the victim client and the server. 
It has full knowledge of the language model architecture and parameters, including the dropout rate and dropout layer positions, as they are part of the model architecture. The objective of the adversary is to recover the private training data of the victim client by utilizing the victim client's shared gradient updates. 
The threat model aligns with a real-world training scenario, where the only communication between the client and the server is the gradient updates. The adversary can only utilize the shared gradient updates to attempt the recovery without extra knowledge of the data, \textit{e.g.,} ground truth labels and sequence lengths.

Existing attacks~\cite{dlg, tag, lamp, film} often adopt a threat model that is with extra knowledge on the ground truth labels and the individual sequence lengths in the data batch, as discussed in Section~\ref{section:related_work:assumptions}. 
We highlight that the threat model defined in our work does not assume such knowledge that gives advantages to the recovery, which aligns with a more practical training scenario.

\subsection{Overview of \codename}
\label{sec:approach:overview}

Figure~\ref{fig:overview} shows the overview of \codename.
We highlight that continuous and discrete optimizations collaborate in our approach. 
In continuous optimization, we enhance the established gradient inversion attack framework~\cite{lamp} by incorporating dropout mask learning~\cite{dropout}.
Then, we employ beam search~\cite{beamsearch2} as a search strategy to reorder the recovered tokens in discrete optimization. 

\paragraph{Recovery loss function}
The recovery loss function is a distance function that measures the distance between the dummy gradients and the original gradients, such as the $L_2$ norm distance~\cite{dlg}, the combination of $L_2$ and $L_1$ norm distance~\cite{tag}, and the cosine distance~\cite{ig}. 
We empirically choose the combination of $L_2$ and $L_1$ norm as the distance measurement in this work.
The recovery loss is defined as
\begin{gather}\small
\label{eq:reconstruction_loss}
\begin{aligned}
\mathcal{L}_{r}(\bm{E}, \bm{y}, \bm{\theta}^t, \nabla_{\bm{\theta}^t}\mathcal{L}(\bm{E}^*,\bm{y}^*)) = & \sum_{i=1}^l \norm{\nabla_{\bm{\theta}^t_i}\mathcal{L}(\bm{E}, \bm{y}) - \nabla_{\bm{\theta}^t_i}\mathcal{L}(\bm{E}^*,\bm{y}^*)}_2 \\
& \hspace{-8pt} + \alpha_{L_1} \norm{ \nabla_{\bm{\theta}^t_i}\mathcal{L}(\bm{E}, \bm{y}) - \nabla_{\bm{\theta}^t_i}\mathcal{L}(\bm{E}^*,\bm{y}^*) }_1,
\end{aligned}
\end{gather}
where $l$ is the number of layers in the language model, and $\alpha_{L_1}$ is the pre-defined regularization weighting factor for the $L_1$ norm.

\input{algorithms/main_algo}

\paragraph{Overview of our approach}
The overall workflow of \codename is listed in Algorithm~\ref{alg:main}.
First, the attacker initializes the dummy dropout mask $\bm{\Psi}$ by $\mathrm{Bernoulli}$ distribution with the dropout rate $p$~(line 1).
Next, it initializes a dummy embedding tensor $\bm{E} \in \mathbb{R}^{b \times n \times d}$ for approximating the original embedding $\bm{E}^*$ and the dummy label tensor $\bm{y} \in \mathbb{R}^{b \times c}$.
We follow the initialization approach from LAMP~\cite{lamp} to initialize $n_{e}$ random embedding candidates and select the one $\bm{E}_{d}$ that yields the minimum recovery loss by Formula~\ref{eq:reconstruction_loss}~(line 2). 
After the initialization, it iteratively executes the continuous and discrete optimizations $n_{h}$ times~(lines 3--8).
Specifically, it employs the continuous optimization contOPT first~(line 4) and then tokenizes the embedding tensor $\bm{E}_{c}$ produced to obtain the recovered tokens $\bm{X}_{c}$~(line 5).
Subsequently, the discrete optimization discOPT takes $\bm{X}_{c}$ as input and outputs the optimized tokens $\bm{X}_{d}$~(line 6).
$\bm{X}_{d}$ is embedded to obtain $\bm{E}_{d}$ and will be used in the next iteration~(line 7).
After iterating all hybrid optimization rounds, we compare the results of continuous and discrete optimizations~(lines 9--12) and output the corresponding recovered tokens.

A running example of \codename in our experiment is shown in Figure~\ref{fig:overview}.
By iteratively alternating the two optimization processes, all individual tokens are recovered and most of the tokens are rearranged in the correct order.

\input{algorithms/continuous_opt}
\input{algorithms/discrete_opt}

\subsection{Continuous Optimization}
\label{sec:approach:continuous_opt}

The main process of our continuous optimization is shown in Algorithm~\ref{alg:continuous_optimization}. 
First, the adversary permutes the dummy embedding to get $n_{p}$ candidates and selects the one with the minimum recovery loss~(line 1).
Subsequently, it updates the dummy embedding, the dummy labels, and the dropout mask with gradient descent based on the recovery loss function and iterates $n_{c}$ times~(lines 2--7). 
Note that our approach considers the frozen embedding layers and the activated dropout~(line 6) in practical settings.

\paragraph{Resolving frozen embedding layers}
Unlike existing attacks~\cite{decep,film,melis}, our approach does not particularly rely on the gradients of these layers.
We run gradient descent from randomly initialized dummy embedding and exclude these layers. 

\paragraph{Resolving activated dropout}
When dropout is activated, the language model of the attacker also employs dropout~\cite{scheliga2023dropout} when computing the recovery loss.
As described in the threat model in Section~\ref{sec:approach:threat_model}, the attacker has knowledge of the model architecture, meaning that the dropout rate and the positions of the dropout layers are known to the attacker.
When the adversary launches the attack, it activates dropout and initializes the dropout mask $\bm{\Psi}$.\footnote{In the case of unknown dropout rates and unknown positions, the attacker can use the empirically well-established convention of dropout rates (\textit{e.g.,} 0.1) and dropout layer positions (\textit{e.g.,} after the transformer attention/output layer) for initialization.}
It then optimizes the dropout mask $\bm{\Psi}$.

After computing the recovery loss~(line 3), we compute the gradients for the dummy embedding, the dummy labels, and the dropout mask. We optimize them simultaneously in an iterative manner~(lines 4--6). 
The values of the dropout masks are clamped between 0 and 1~(line 6).

\subsection{Discrete Optimization}
\label{sec:approach:discrete_opt}

Algorithm~\ref{alg:discrete_algorithm} presents the main procedure of our discrete optimization method.
The output of continuous optimization is the optimized embedding tensor $\bm{E}_c$. 
It is transformed into individual tokens $\bm{X}_c$ by comparing with the token embeddings of the language model.
Considering that the recovered tokens are usually in an incorrect order~\cite{lamp, dlg, tag}, we propose to utilize beam search~\cite{ beamsearch2} to reorder the recovered tokens due to its systematic search ability. Other alternatives such as Nucleus Sampling~\cite{holtzman2019curious} for generative models may not be directly applicable, as the victim model is a classification model.

At the beginning of the process, permutations and selections are employed on $\bm{X}$~(line 1), which is similar to the operations in continuous optimization.
Note that we keep the first $n_b$ permutations $\{\bm{X}_{1..n_b}\}$ with the minimum recovery loss values as our initialization of the beams.
Then $n_d$ times beam search are employed for each $\bm{X}_j$ with the corresponding token set $\{\bm{T}_j\}$ to update $\bm{X}_j$~(lines 3--7).
Specifically, for each sequence in each beam, we go over each of the positions and substitute it with each of the recovered tokens for that sequence. 
For each position of the sequence, the initial $n_{b}$ candidates with the lowest recovery loss values are retained as the new beams. It then progresses to the subsequent position and the process is reiterated. This is similar to a greedy search approach, with the distinction being the retention of $n_{b}$ candidates as opposed to just the current optimal one.

When the batch size is larger than 1, the shorter sequences in the batch are padded with the padding token to the longest length. This causes extra challenges in recovering the shorter sequences in terms of recovery and efficiency. To overcome these difficulties, we include the padding token in the token set $\{\bm{T}_j\}$ in the reordering. When the padding token is selected at a position, it injects a crucial ending signal, which helps prevent shorter sequences from appending unnecessary tokens to improve the recovery.

%% file: algorithms/main_algo.tex
\begin{algorithm}[t]

\caption{\codename - Main Algorithm}
\label{alg:main}
\begin{flushleft}
\textbf{Input:} 
$\bm{\theta}^t$ - model parameters at $t$-th training iteration; \\
$\nabla_{\bm{\theta}^t}\mathcal{L}(\bm{E}^*,\bm{y}^*)$ - original gradients at $t$-th training iteration;\\
\textbf{Output:} $\bm{X}$ - optimized token solution;
\end{flushleft}
\begin{algorithmic}[1]

\STATE $\bm{\Psi} \gets \mathrm{Bernoulli}(p);$
\STATE $\bm{E}_{d}, \bm{y} \gets \mathrm{initDummyData}(n_{e}, \bm{\theta}^t, \nabla_{\bm{\theta}^t}\mathcal{L}(\bm{E}^*,\bm{y}^*))$;

\FOR{$i \in 0..n_{h}$}
\STATE $\bm{E}_{c}, \bm{y} \gets \mathrm{contOpt}(\bm{E}_{d}, \bm{y}, \bm{\theta}^t, \nabla_{\bm{\theta}^t}\mathcal{L}(\bm{E}^*,\bm{y}^*), \bm{\Psi});$
\STATE $\bm{X}_{c} \gets \mathrm{tokenize}(\bm{E}_{c});$
\STATE $\bm{X}_{d} \gets \mathrm{discOpt}(\bm{X}_{c}, \bm{y}, \bm{\theta}^t, \nabla_{\bm{\theta}^t}\mathcal{L}(\bm{E}^*,\bm{y}^*));$
\STATE $\bm{E}_{d} \gets \mathrm{embed}(\bm{X}_{d});$
\ENDFOR
\STATE $\bm{X} \gets \bm{X}_{d};$
\IF{$\mathcal{L}_{r}(\bm{E}_{c}, \bm{y}, \bm{\theta}^t, \nabla_{\bm{\theta}^t}\mathcal{L}(\bm{E}^*,\bm{y}^*)) > \mathcal{L}_{r}(\bm{E}_{d}, \bm{y}, \bm{\theta}^t, \nabla_{\bm{\theta}^t}\mathcal{L}(\bm{E}^*,\bm{y}^*))$}
\STATE $\bm{X} \gets \bm{X}_{c};$
\ENDIF
\RETURN $\bm{X}$

\end{algorithmic}
\end{algorithm}

%% file: algorithms/continuous_opt.tex
\begin{algorithm}[t]

\caption{Continuous Optimization - contOpt($\cdot$)}
\label{alg:continuous_optimization}
\begin{flushleft}
\textbf{Input:} 
$\bm{E}$ - dummy embedding; \\
$\bm{y}$ - dummy labels; \\
$\bm{\theta}^t$ - model parameters at $t$-th training iteration; \\
$\nabla_{\bm{\theta}^t}\mathcal{L}(\bm{E}^*,\bm{y}^*)$ - original gradients at $t$-th training iteration;\\
$\bm{\Psi}$ - dropout mask of attacker model; \\
\textbf{Output:} $\bm{E}$ - optimized dummy embedding;
\end{flushleft}
\begin{algorithmic}[1]

\STATE $\bm{E} \gets \tightunderset{\tiny{\quad\bm{E}' \in \mathrm{permute}(\bm{E}, n_{p})}}{\arg\min} \mathcal{L}_{r}(\bm{E}', \bm{y}, \bm{\theta}^t, \nabla_{\bm{\theta}^t}\mathcal{L}(\bm{E}^*,\bm{y}^*))$;
\FOR{$i \in 0$..$n_{c}$}
\STATE $L_{r} \gets \mathcal{L}_{r}(\bm{E}, \bm{y}, \bm{\theta}^t, \nabla_{\bm{\theta}^t}\mathcal{L}(\bm{E}^*,\bm{y}^*));$
\STATE $\bm{E} \gets \bm{E} - \lambda \cdot \frac{\partial{L_{r}}}{\partial{\bm{E}}};$
\STATE $\bm{y} \gets \bm{y} - \lambda \cdot \frac{\partial{L_{r}}}{\partial{\bm{y}}};$
\STATE $\bm{\Psi} \gets \min(1, \max(0, \bm{\Psi} - \lambda \cdot \frac{\partial{L_{r}}}{\partial{\bm{\Psi}}}));$
\ENDFOR
\RETURN $\bm{E}$

\end{algorithmic}
\end{algorithm}

%% file: algorithms/discrete_opt.tex
\begin{algorithm}[t]

\caption{Discrete Optimization - discOpt($\cdot$)}
\label{alg:discrete_algorithm}
\begin{flushleft}
\textbf{Input:} 
$\bm{X}$ - dummy sequences of tokens; \\
$\bm{y}$ - dummy labels; \\
$\bm{\theta}^t$ - model parameters at $t$-th training iteration; \\
$\nabla_{\bm{\theta}^t}\mathcal{L}(\bm{E}^*,\bm{y}^*)$ - original gradients at $t$-th training iteration;\\
\textbf{Output:} $\bm{X}$ - optimized dummy sequences of tokens;
\end{flushleft}
\begin{algorithmic}[1]

\STATE $\{\bm{X}_{1..n_b}\} \gets \tightunderset{\tiny{\qquad\bm{X}' \in \mathrm{permute}(\bm{X}, n_{p})}}{\arg\min} \left( \mathcal{L}_{r}(\bm{X}', \bm{y}, \bm{\theta}^t, \nabla_{\bm{\theta}^t}\mathcal{L}(\bm{E}^*,\bm{y}^*)),n_b \right)$;
\STATE $\{\bm{T}_{1..n_b}\} \gets \mathrm{getTokens}(\{\bm{X}_{1..n_b}\})$
\FOR{$i \in 0$..$n_{d}$}
\FOR{$j \in 0$..$n_{b}$}
\STATE $\bm{X}_{j} \gets \tightunderset{\tiny{\qquad\bm{X}'_j \in \mathrm{beamSearch}}(\bm{X}_j, \{\bm{T}_j\})}{\arg\min} \mathcal{L}_{r}(\bm{X}'_j, \bm{y}, \bm{\theta}^t, \nabla_{\bm{\theta}^t}\mathcal{L}(\bm{E}^*,\bm{y}^*))$;
\ENDFOR
\ENDFOR
\STATE $\bm{X} \gets \tightunderset{\tiny{\quad\bm{X} \in \{\bm{X}_{1..n_b}\}}}{\argmin}\mathcal{L}_{r}(\bm{X}, \bm{y}, \bm{\theta}^t, \nabla_{\bm{\theta}^t}\mathcal{L}(\bm{E}^*,\bm{y}^*))$;
\RETURN $\bm{X}$

\end{algorithmic}
\end{algorithm}

%% file: sections/5_experiment_cr.tex
\section{Experimental Evaluation} \label{sec:evaluation}
This section evaluates \codename from the following four key aspects.

\paragraph{Benchmark performance}
Following the experimental setup in~\cite{lamp}, we benchmark the performance of \codename against all baseline approaches~\cite{dlg, tag, lamp} in benchmark settings. 

\paragraph{Practical performance}
We assess the practicality of all approaches by evaluating their performance in practical settings, which are more constrained.

\paragraph{Ablation studies}
We conduct ablation studies to evaluate the performance of \codename under different circumstances, \textit{i.e.,} various model sizes/types and various dropout rates in practical settings. Additionally, to further evaluate the practicality of \codename in the absence of the assumptions discussed in Section~\ref{section:related_work:assumptions}, we explore instances where we relax one or both of these premises.

\paragraph{Resilience against proactive defenses} To evaluate the resilience of \codename, we compare the performance of all approaches against proactive defenses in practical settings, \textit{i.e.,} gradient noise~\cite{dlg, wei2020framework} and gradient pruning~\cite{dlg}.

Table~\ref{table:example} shows a recovery example of all approaches on the CoLA dataset with a batch size of 2 on the $\text{BERT}_{base}$ model in practical settings, where correctly recovered unigrams and longer sequences are highlighted in yellow and green. Note that \codename recovers significantly more bigrams and longer sequences.
\input{tables/example}

\subsection{Experimental Settings}
\label{sec:evaluation:settings}

In this section, we introduce the experimental settings, covering training settings, datasets, language models, baselines, hyperparameters, FL setup, metrics, and result statistics. 
Further details on additional experimental settings, including empirical experiment implementation, model implementation in practical settings, and detailed metric calculation are provided in Appendix~\ref{appendix_B}.

\paragraph{Training settings} We evaluate \codename in two different training settings, namely \emph{benchmark} and \emph{practical settings}. In benchmark settings, 
following existing attacks~\cite{dlg, tag, lamp}, the embedding layers are trainable and observable, but the attacker does not directly extract tokens from the gradients of these layers. 
Additionally, the dropout layers are deactivated during training. In practical settings, the two lightweight defense mechanisms discussed in Section~\ref{sec:related_work:defense} are applied. In particular, the embedding layers are frozen, such that there are no gradient updates to be observed for these layers. Additionally, 
dropout is activated during training to prevent overfitting. 
In our comparative study in Section~\ref{sec:evaluation:benchmark} and Section~\ref{sec:evaluation:practicality}, both settings adopt the known label assumption and the known sequence lengths assumption for comparison with baseline methods. In the ablation study in Section~\ref{sec: assumptions}, we relax these assumptions and show that \codename does not need to operate under such assumptions.

\paragraph{Datasets} 
Our main experiments are conducted on three widely used binary sequence classification datasets, CoLA~\cite{cola}, SST-2~\cite{sst2}, and Rotten Tomatoes~\cite{rm}, with typical sequence lengths of 5--9 words, 3--13 words, and 14--27 words, respectively. 
The selected datasets are representative in terms of sequence lengths, as other commonly used datasets such as Tweet Sentiment Extraction~\cite{tweeter} (5--25 words) and Yahoo Answers Topics~\cite{yahoo} (5--15 words) have similar sequence lengths. 
To ensure a fair comparison, for each dataset, we randomly select 64 samples with the consistent random seed from the implementation of LAMP~\cite{lamp}. Later in Section~\ref{sec:multiclass}, we evaluate \codename with a 10-class classification task with the Yahoo Answers Topics dataset.

\paragraph{Language models} 
The main experiments (in Section~\ref{sec:evaluation:benchmark} and \ref{sec:evaluation:practicality}) are conducted on the most extensively used language model, $\text{BERT}_{base}$ model~\cite{bert}, with 12 layers, 12 heads, and 110 million parameters. 
Later in Section \ref{sec:evaluation:various_models}, models with various sizes/types are discussed and evaluated. 

\paragraph{Baselines}
Three existing approaches, DLG~\cite{dlg}, TAG~\cite{tag}, and LAMP~\cite{lamp} are used as the baselines.
In particular, two variants, $\text{LAMP}_{cos}$~(using cosine distance) and $\text{LAMP}_{L_2+L_1}$~(using combined $L_2$ and $L_1$ distance) of LAMP are used.
Implementations of all baseline approaches are taken from the released source code of LAMP. 
We adopt their by-default hyperparameter settings in the evaluation. 
The benchmark settings we define are the same as those used in the baseline methods. 
In practical settings, our configurations are designed to mimic realistic conditions, such as frozen embedding and activated dropout. Since the hyperparameters (\textit{e.g.,} number of iterations) are not related to these specific configurations, we keep the same hyperparameter settings.

\paragraph{Experimental parameters}
We empirically set the number of embedding candidates $n_e = 2,000$ and the number of permutations $n_p = 2,000$ to obtain an initial solution, considering the sensitivity of continuous optimization.
For the optimizations, we set the iteration number of hybrid optimization $n_h=5$, iteration number of continuous optimization $n_c=2,000$, iteration number of discrete optimization $n_d=5$, and number of beams $n_b = 4$, reaching a total of $10,000$ continuous optimization steps and $25$ discrete optimization steps, which empirically achieve a reasonable trade-off considering efficiency and efficacy.
Additionally, we set the $L_1$ weight factor $\alpha_{L_1} = 0.01$, and learning rate $\lambda = 0.01$ with a learning rate decay factor $\gamma = 0.89$ for every 50 steps, which aligns with $\text{LAMP}_{L_2+L_1}$.
We use the AdamW optimizer~\cite{adamw}.
Early stopping occurs when the discrete optimization does not improve the output of the continuous optimization. All these parameters have been kept consistent in both baseline settings and practical settings for a fair comparison with baseline methods.

We note that smaller iteration numbers for continuous optimization are used in the default settings of DLG~(2,500), TAG~(2,500), and LAMP~(2,000).
To ensure a fair comparison, we standardize the iteration count to 10,000 across all approaches in a trial experiment.
Our approach consistently shows improvement, whereas other approaches early fall into a local optimum at around their default number of iterations.
For this reason, we keep the default number of iterations for all baselines.

\paragraph{FL setup} Given the conceptual similarity between the setups with multiple clients and a single client-server pair in FL~\cite{film}, this work utilizes the configuration of one server and one client (victim) for simplicity, which is commonly adopted in prior studies~\cite{dlg, tag, lamp, film}. Specifically, a scenario with $m$ clients, each possessing $b$ samples, is functionally equivalent to training a model with a batch size of $m \times b$~\cite{film}. We empirically demonstrate this equivalence in Table~\ref{table:multi} in Appendix~\ref{FL_setup}, where the performance of our approach under both setups is consistent.

\paragraph{Metrics}
The efficacy of the attacks is evaluated by the recovery rate of the original data batch.
We use the ROUGE score~\cite{rouge} as a metric, aligning with existing work~\cite{lamp,film}.
The ROUGE score measures the similarity between the original sentences and the recovered sentences.
Specifically, different overlapping grams are used, including unigram, bigram, and longest-matching subsequences, denoted by ROUGE-1~(R-1), ROUGE-2~(R-2), and ROUGE-L~(R-L). The aggregated F-score is used to compute the ROUGE score.
When computing the ROUGE score, special tokens without meaningful semantic~(start token [CLS], end token [SEP], and padding token [PAD]) are excluded.
In ablation studies and the proactive defenses evaluation, we use Matthews Correlation Coefficient (MCC)~\cite{mcc} to measure the utility of the models.

During our benchmarking in Section~\ref{sec:evaluation:benchmark} and Section~\ref{sec:evaluation:practicality}, the original training samples~(\textit{i.e.,} the ground truth) are used to evaluate \codename's capability of recovering text that closely matches the actual training data, in terms of tokens and token order.
However, in real-world scenarios, the original data is unknown to the attacker. 
The attacker can only infer that the recovered data contains matched tokens and token order to an extent based on the exhibited close gradient distances.

\paragraph{Result statistics}
As random factors such as the initialization of dropout masks and dummy embedding can impact the final recovery results, we report the aggregated results of three random runs for a more reliable evaluation.
Specifically, mean values and standard deviations of R-1, R-2, and R-L with different batch sizes in powers of 2 (up to 32) are presented to evaluate the performance of different approaches.

\subsection{Benchmark Evaluation}
\label{sec:evaluation:benchmark}

We benchmark the performance of \codename with all baseline approaches with batch sizes from 1 to 32 in benchmark settings.
The results are shown in Figure~\ref{fig:benchmark_setting_performance}. 
\input{figures/benchmark_performance}
\input{figures/versatile_performance}

\codename recovers a significant portion of the data, with up to 92.9\% recovery rate, demonstrating its remarkable recovery ability. Despite some occasional larger standard deviations, it outperforms all baseline approaches across all datasets and batch sizes, with improvements of 13.2\% in R-1, 15.5\% in R-2, and 13.6\% in R-L on average, and a maximum improvement of 28.9\% compared to the best baseline approach.
The results indicate the superior recovery ability of \codename in benchmark settings. Additionally, it also showcases its better generalizability with influential random factors (\textit{e.g.,} dummy embedding initialization).
We also observe a general trend where performance decreases as batch size increases. This is due to a larger batch size resulting in more averaged gradients of the sequences, which poses challenges to the recovery process.

\codename not only recovers more individual tokens (R-1) but also rearranges more tokens in a meaningful and sensible order (R-2 and R-L), making the recovery leak more important semantics. In benchmark settings, this notable improvement is mainly achieved by the effective beam search technique, compared to no discrete optimization at all (DLG and TAG) or only random search (LAMP).

\subsection{Practical Evaluation}
\label{sec:evaluation:practicality}
We evaluate the practicality of \codename and the baseline approaches by shifting from benchmark settings to practical settings. This involves applying frozen embedding layers and activated dropout, which not only brings inherent challenges to cope with but also better represents real-world scenarios. We run all approaches with batch sizes ranging from 1 to 32. Note that two variants of \codename are assessed, \textit{i.e.,} with or without dropout mask learning. The results are shown in Figure~\ref{fig:versatile_setting_performance}.

\input{tables/RQ3_1}

We observe that the performance of all baseline approaches and \codename without dropout mask learning decreases in practical settings compared to the performance in benchmark settings. This performance drop affirms that the frozen embedding layers and activated dropout indeed pose difficulty to the gradient matching. Despite this, \codename without dropout mask learning still recovers a large portion of the data with up to 76.9\% recovery rate. Our recovery is more coherent in comparison to existing approaches. It outperforms the baseline approaches in most cases, with improvements of 13.2\% in R-1, 10.5\% in R-2, and 11.8\% in R-L on average and 28.7\% maximum compared to the best baseline approach with some occasional larger standard deviations. Such results underscore the efficacy and generalizability of our approach in practical settings even in the absence of the dropout mask learning strategy.

\input{figures/dropout_performance}

\codename with the dropout mask learning technique demonstrates an even more remarkable recovery ability in practical settings.
It remains consistent performance in most cases compared to performance in benchmark settings, affirming that the dropout mask learning technique can indeed offset the dropout-induced noise. 
\codename with dropout mask learning outperforms all baseline approaches on all datasets with all batch sizes, with remarkable improvements of 31.8\% in R-1, 25.5\% in R-2, and 26.8\% in R-L on average and 48.5\% maximum, demonstrating significant superiority and reaffirming the effectiveness of the dropout mask learning technique.

\subsection{Ablation Studies}
\label{sec:evaluation:ablation}

We evaluate the effectiveness of \codename under different circumstances, \textit{i.e.,} various model sizes/types, and various dropout rates. 
Subsequently, we relax the two assumptions, \textit{i.e.,} known labels and known exact sequence lengths, to evaluate the impact of them. In addition, we also evaluate the generalizability of \codename with a multi-class classification task and with larger batch sizes.
In most cases, if not otherwise noted, all approaches are run with their default hyperparameters setting and on the CoLA dataset with a batch size of 1 in practical settings.

\input{tables/RQ3_5_7}

\subsubsection{Various Model Sizes and Types}

\label{sec:evaluation:various_models}

Different sizes of BERT models are chosen for comparison.
Besides $\text{BERT}_{base}$, another two structured models of the BERT model are chosen, which are $\text{BERT}_{tiny}$~\cite{tinybert} and $\text{BERT}_{large}$~\cite{bert}.
$\text{BERT}_{tiny}$ has 6 layers, 12 heads, and 67 million parameters.
$\text{BERT}_{large}$ has 24 layers, 16 heads, and 336 million parameters.
We compare both variants of \codename and all baseline approaches on these models.

To evaluate \codename on various model types, we select RoBERTa~\cite{roberta}, which is a variant of BERT. 
Three sizes are chosen, which are $\text{RoBERTa}_{tiny}$~\cite{robertatiny} with 4 layers, 8 heads, and 28 million parameters, $\text{RoBERTa}_{base}$ with 12 layers, 12 heads and 125 million parameters, and $\text{RoBERTa}_{large}$, with 24 layers, 16 heads and 355 million parameters.
As the baseline approaches do not support attacking RoBERTa models in their implementations, we only evaluate our approach. 

The results on various BERT models are shown in Table~\ref{table:RQ3_1_1}.
For $\text{BERT}_{tiny}$ and $\text{BERT}_{base}$, as the model size increases, the performance of both variants of \codename and LAMP decreases as larger models have more parameters, posing difficulty to the optimization.
However, the performance of DLG and TAG increases as the model size increases, aligning with the observation from TAG~\cite{tag}.
This is due to the difference in methodologies, as they only employ continuous optimization.
Smaller models contain less information for these approaches to reach an optimum.
However, this phenomenon is only observed in $\text{BERT}_{tiny}$ and $\text{BERT}_{base}$.
In $\text{BERT}_{large}$, which contains more layers and parameters, the performance of all approaches decreases dramatically except for \codename, which utilizes dropout learning. This underscores the effectiveness of our dropout mask learning technique in countering the negative impacts of dropout, as a larger model incorporates more dropout layers.

Note that there is another hyperparameter setting for baseline approaches proposed by Balunovic \textit{et al.}~\cite{lamp} specifically for $\text{BERT}_{large}$ to counter the optimization difficulty posed by the architectures and parameters.
This setting involves several changes, \textit{i.e.,} more optimization steps, a different optimizer, a different learning rate decay scheduler, and a gradient clipping technique.
We acknowledge that this setting can indeed resolve the optimization difficulty, and we adapt part of it into our approach.
Specifically, a linear learning rate decay scheduler is adopted and the gradient clipping technique~\cite{breaching} is applied to clip the magnitude of the dummy embedding's gradient to 0.5 for more stable optimization, while the rest of the hyperparameters setting remains the same.
All approaches are evaluated with the tuned hyperparameters setting and the results are shown in Table~\ref{table:RQ3_1_3}.
The performance of all approaches is boosted significantly, and our approach with dropout mask learning still outperforms all baseline approaches.

For RoBERTa models, we first run our approach on all three sizes with the default hyperparameter setting, and the results are shown in Table~\ref{table:RQ3_1_1}.
The performance of \codename follows the same trend as in BERT models, the attack is weakened when model size increases.
For $\text{RoBERTa}_{large}$, our attack without the incorporation of dropout mask learning fails to recover any meaningful text, while the method that incorporates dropout mask learning is only partially successful.
The overall performance of \codename on RoBERTa models falls short compared to BERT models, attributable to RoBERTa's increased complexity and more parameters.

The tuned hyperparameter setting is applied on $\text{RoBERTa}_{large}$, and the results are shown in Table~\ref{table:RQ3_1_3}.
We observe that the performance of \codename with dropout mask learning is boosted significantly, while the performance of \codename without dropout mask learning only increases by a small level. As the $\text{RoBERTa}_{large}$ has more dropout layers, this further affirms the efficacy of applying the dropout mask learning technique to offset the dropout-induced noise.

We emphasize that \codename has a greater generalization ability to different model types, compared to existing attack LAMP~\cite{lamp}, as LAMP involves a customized external auxiliary language model (GPT-2) which employs the same tokenizer as BERT for calculating perplexity~\cite{perplexity} as part of the recovery loss.
In the current implementation of LAMP, it does not provide the option to attack different model types.
It is unclear how abandoning the auxiliary language model would affect the recovery.

The results of this ablation study demonstrate that \codename remains effective with various model sizes/types, revealing significant privacy vulnerabilities of language models in general.

\subsubsection{Varying Dropout Rates}
\label{sec:evaluation:dropout_rates}

As mentioned in Section~\ref{section:settings} and Section~\ref{sec:related_work:defense}, dropout serves as a technique to prevent overfitting in model training and also a lightweight defense mechanism against gradient inversion~\cite{ scheliga2023dropout}.
To assess \codename under the scenarios with different dropout rates, we evaluate the performance of all approaches on $\text{BERT}_{base}$ model with increasing dropout rates.
As a higher dropout rate simultaneously worsens the model utility, following the practice from LAMP, we train the models with 2 epochs and report the utility metric MCC~\cite{mcc}. 
We observe that with a dropout rate from 0.1 to 0.4, the MCC drops from 0.77 to 0, indicating that the model is completely dysfunctional. 
Thus, we do not evaluate a dropout rate higher than this.

Figure~\ref{fig:dropout_performance} shows the stable superiority of \codename with dropout mask learning compared to all baselines, despite the decreasing performance of all approaches.
While the attack is weakened as the dropout rate increases, the utility of the models simultaneously has a dramatic deterioration, which is unacceptable in practical cases.
\codename can still recover a large portion of the data even if the models are dysfunctional in terms of utility. 
This indicates that \codename does not rely on low dropout rates.
Hence, increasing dropout rates cannot serve as an effective defense mechanism against \codename.

\subsubsection{Assumptions Relaxation}
\label{sec: assumptions}

We further evaluate the impact of the two adopted assumptions by existing work discussed in Section~\ref{section:related_work:assumptions}, \textit{i.e.,} known labels and known sequence lengths. 

\input{figures/noise_performance}

\paragraph{Without known labels assumption}
To launch the attack without known labels assumption, the attacker randomly initializes dummy labels and optimizes them simultaneously with the dummy embedding and the dropout mask.
We run \codename with dropout mask learning without the known labels assumption on the CoLA dataset with batch sizes from 1 to 32 on $\text{BERT}_{base}$ model.
The results are shown in Table~\ref{table:RQ3_5_7}.
We observe that the performance of \codename remains consistent, compared to when the assumption exists, indicating that \codename is insensitive to this assumption.

\paragraph{Without known exact sequence lengths assumption}
To launch the attack without the known exact sequence lengths assumption, we assume that only the longest length in the batch is known to the attacker, as it is significantly more practical to enumerate all possible longest lengths for the whole batch, compared to the existing assumption~\cite{lamp} where the attacker needs to enumerate all possible lengths for each sequence in the batch. This existing assumption becomes impractical when batch size grows.
We run \codename with dropout mask learning without the known exact sequence lengths assumption on the CoLA dataset with batch sizes from 2 to 32 on $\text{BERT}_{base}$ model, as batch size 1 only involves a single sentence in the batch.
The results are shown in Table~\ref{table:RQ3_5_7}. \codename achieves an even higher attack efficacy, until the batch size of 16.
This is expected, as the recovered sequence lengths are no longer constrained if the sequences are recovered at different indexes, and by involving the padding token, the length for each index can be adjusted dynamically towards a lower recovery loss.
The shorter sequences can append padding tokens and the longer sequences do not get cut off.
The results indicate that \codename is also insensitive to this assumption.

\paragraph{Without both assumptions}
We run \codename with dropout mask learning without both assumptions on the CoLA dataset with batch sizes from 2 to 32 on $\text{BERT}_{base}$ model, and the results are shown in Table~\ref{table:RQ3_5_7}. We observe a consistent attack efficacy phenomenon, further indicating that \codename is insensitive to both assumptions and can even perform better without these assumptions, mainly because the sequence lengths constraint is lifted.

The results of this ablation study further showcase the practicality of \codename in the absence of the two assumptions, where the prior knowledge of the labels and the exact sequence lengths is not provided to the attacker.

\subsubsection{Multi-class Classification}
\label{sec:multiclass}

The evaluation of \codename is with a binary classification task. 
When the known label assumption is relaxed, 
the difference between multi-class and binary classification mainly lies in the possible number of ground truth labels.
To assess the generalization of \codename to such a more challenging task, we evaluate it with a 10-class classification task with the Yahoo Answers Topics dataset~\cite{yahoo} under practical settings without the known label assumption.
The same data selection procedure as our benchmark datasets is followed. 
The results are shown in Table~\ref{table:multi-class} in Appendix~\ref{appendix_A}. 
The evaluation demonstrates that \codename is resilient to such challenging tasks, and can effectively recover both the ground truth labels and the textual data, regardless of the number of possible ground truth labels.

\subsubsection{Large Batch Sizes}
\label{sec:worst-case}
As shown in Figure~\ref{fig:benchmark_setting_performance} and Figure~\ref{fig:versatile_setting_performance}, the attack efficacy for all approaches wanes as the batch size grows, mainly due to the gradient information for all sequences in the batch being averaged together. 
To further explore the effect of the batch size, we evaluate \codename with the CoLA dataset with larger batch sizes up to 128~\footnote{Extra data is selected with the same data selection procedure as in the original data selection with the same random seed.}, the largest batch size used originally to pre-train BERT~\cite{bert-pretrain}. The results are shown in Table~\ref{table:large-b-size} in Appendix~\ref{appendix_A}. 
The evaluation demonstrates that the attack efficacy of \codename continues to drop as batch size grows. However, a significant portion of the private data is still recovered, further showcasing the capability of \codename and underlying the severity of this privacy threat, even with large batch sizes.

\subsection{Proactive Defenses}
\label{sec:evaluation:defenses}

We consider two widely used proactive defenses, \textit{i.e.,} gradient noise~\cite{wei2020framework, dlg} and gradient pruning~\cite{dlg}, which aim to counteract gradient inversion. 
We explore their effect in all approaches in practical settings on $\text{BERT}_{base}$ model. Huang \textit{et al.}~\cite{huang2021evaluating} evaluate other defense mechanisms such as gradient encryption~\cite{aono2017privacy} and input encoding~\cite{zhang2017mixup, huang2020instahide}. Since gradient encryption requires a special setup and is costly to run, and input encoding is primarily for image data, we do not evaluate them in this work.

\input{tables/noise-tradeoff}

\subsubsection{Gradient Noise}

\input{tables/RQ4_2}
\noindent Zhu \textit{et al.}~\cite{dlg} and Wei \textit{et al.}~\cite{wei2020framework} suggest a defense that adds Gaussian or Laplacian noise to the gradients.
We follow a standard DP-SGD~\cite{dp} process, where the client normalizes/clips the gradients before adding the noise.
However, if gradient noise is not a part of the FL training protocol, this can hint that it is used as a defense.
The attacker can check the use of gradient noise by passing a dummy data sample through the model and monitoring the differences in the norm magnitudes of the dummy gradients and the client's gradients. 
If the difference exceeds a threshold, the attacker can deduce the defense is in place and adjust the magnitude of the dummy gradients for a more effective search.

To evaluate the privacy-utility trade-off, the language model is trained on the full CoLA training set (8,551 samples) for 2 epochs with a range of noise levels from 0.001 to 0.05. The noise levels, model utility (MCC), and privacy budgets ($\epsilon$) are summarized in Table~\ref{table:noise-tradeoff}. A default target delta  ($\delta$) of $1 \times 10^{-5}$ is used when computing the privacy budgets. Noise levels beyond 0.05 are not explored as the MCC of the model drops significantly from 0.77 to 0.25. We observe that the model is sensitive to small noise levels in terms of model utility, following similar observations in previous works~\cite{lamp}. Specifically, as language models usually contain a large number of parameters, they are generally pre-trained on a large corpus, so the gradients obtained by the attacker with training on downstream tasks tend to be small. Clipping and adding noise leads to a significant model utility loss, even with only small noise levels and impractically large privacy budgets measured with DP.

To evaluate the attack efficacy against the gradient noise defense, we consider four variants of \codename, which include combinations of either enabling or disabling dropout mask learning or norm normalization.
Figure~\ref{fig:noise_performance} demonstrates that \codename, which combines dropout mask learning and norm normalization, outperforms all baseline approaches, achieving substantial data recovery across almost all noise levels and demonstrating the enhancement and effectiveness of both techniques. Similar to dropout, gradient noise weakens the attack, but higher noise levels sacrifice the model utility. At the noise level of 0.05, the model is nearly dysfunctional in terms of model utility, and \codename with only norm normalization actually outperforms \codename with both techniques, as the noise level is high enough to perturb the dropout mask learning process. Nevertheless, gradient noise as a defense mechanism cannot mitigate \codename, even by compromising most of the model utility.

\subsubsection{Gradient Pruning}
Gradient pruning is another defense by modifying the shared gradients in FL~\cite{dlg}. 
We follow the gradient pruning variant evaluated in LAMP~\cite{lamp} to prune a certain percentage of the gradients. 
The attacker can infer whether this defense is employed by observing the zero entries of the client's gradients. 
Capitalizing on this knowledge, the corresponding gradient entries in the dummy gradients of the attacker can also be zeroed out to aid the gradient matching process for improved data recovery. 

The language model is trained on the full CoLA dataset for 2 epochs with pruning percentages from 0.75 to 0.99. 
We run experiments on 4 variants of \codename, which are combinations of turning on and off dropout mask learning or pruning mask. Table~\ref{table:RQ4_2} shows \codename with both techniques combined recovers a significant portion of the data and remains consistent across all prune percentages, while all baseline approaches have a significant drop in recovery rate starting from prune percentage 0.75, or even completely mitigated. This deviates from the observation of LAMP, as their experiments are run in benchmark settings. These results demonstrate that gradient pruning cannot mitigate our attack, whereas it successfully mitigates all other approaches in practical settings.

%% file: tables/example.tex
\begin{table}[t!]
\definecolor{refcolor}{rgb}{0.15, 0.38, 0.61}
\definecolor{bigramcolor}{rgb}{0.03, 0.47, 0.19}
\definecolor{unigramcolor}{rgb}{0.85, 0.57, 0.0}
\definecolor{bigramcolor}{rgb}{0.7, 0.93, 0.36}
\definecolor{unigramcolor}{rgb}{0.99, 0.97, 0.37}   

\newcommand{\unigram}[1]{\colorbox{unigramcolor}{\vrule height4.2pt depth1pt width0pt #1}}
\newcommand{\bigram}[1]{\colorbox{bigramcolor}{\vrule height4.2pt depth1pt width0pt #1}}
\centering  
\scriptsize
\caption{Text recoverability across different approaches.} 

\label{table:example}

\renewcommand{\arraystretch}{1}

\setlength{\tabcolsep}{3pt}
\begin{tabular}{lll}
\toprule
\multirow{1}{*}{\textcolor{refcolor}{\textbf{Reference}}} & 
\textbf{\textcolor{refcolor}{was the child running to the car?}} & 
\textbf{\textcolor{refcolor}{mary is shorter than five feet.}} \\
\midrule

\multirow{1}{*}{DLG} &
"?.. \unigram{child} \unigram{is} \unigram{was} &
broken.meries youth area finals \\ 
\midrule

\multirow{1}{*}{TAG} & 
are derrick \unigram{the}? \unigram{mary} \unigram{child} toward &
\unigram{feet} \unigram{than}.eth \bigram{mary is shorter} \\ 
\midrule

\multirow{1}{*}{LAMP$_{cos}$} & 
\bigram{mary is} trailer? inventor \bigram{the car}? &
\bigram{shorter than} \bigram{mary is}? \unigram{mary}\\ 
\midrule

\multirow{1}{*}{LAMP$_{L_1L_2}$} & 
\unigram{to} \unigram{child} \unigram{was} \unigram{running} \bigram{the car}? &
feetlle \unigram{mary} \unigram{shorter} \unigram{is} \unigram{feet}. \\
\midrule

\multirow{1}{*}{\codename} & 
\unigram{was} \bigram{running to the car} \bigram{the child}? &
\bigram{mary is shorter than five feet}. \\ 
\bottomrule

\end{tabular}

\end{table}

%% file: figures/benchmark_performance.tex
\begin{figure*}[h!]
    \centering
    
    \begin{subfigure}[b]{0.017\linewidth}
        \raisebox{1.5\height}{\includegraphics[width=\linewidth]{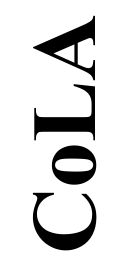}}
    \end{subfigure}
    \hfill
    \begin{subfigure}[b]{0.32\textwidth}
        \includegraphics[width=\textwidth]{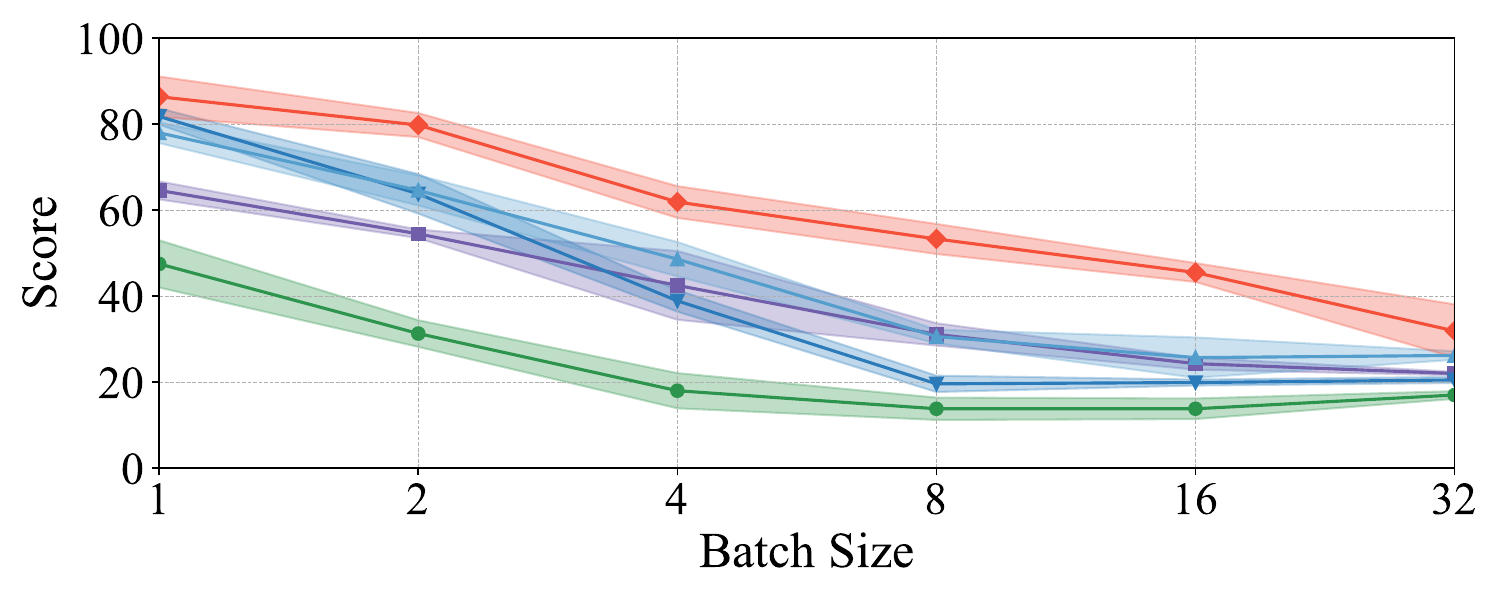}
    \end{subfigure}
    \hfill
    \begin{subfigure}[b]{0.32\textwidth}
        \includegraphics[width=\textwidth]{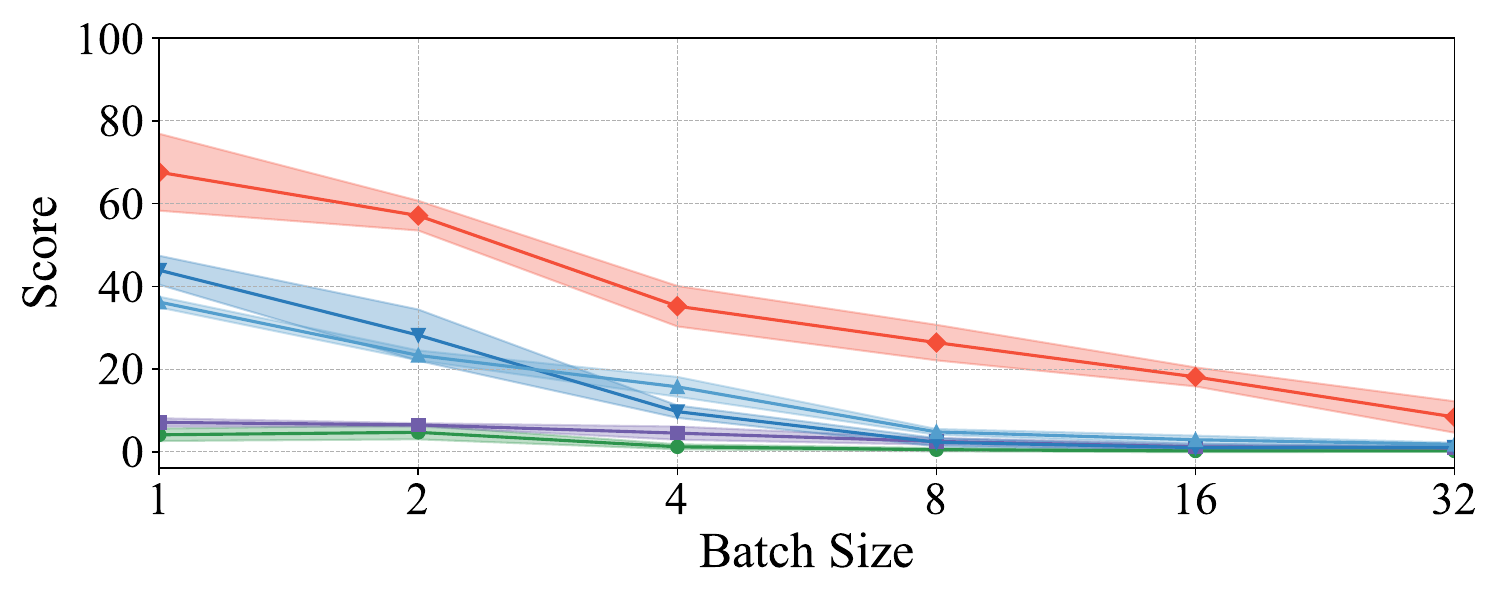}
    \end{subfigure}
    \hfill
    \begin{subfigure}[b]{0.32\textwidth}
        \includegraphics[width=\textwidth]{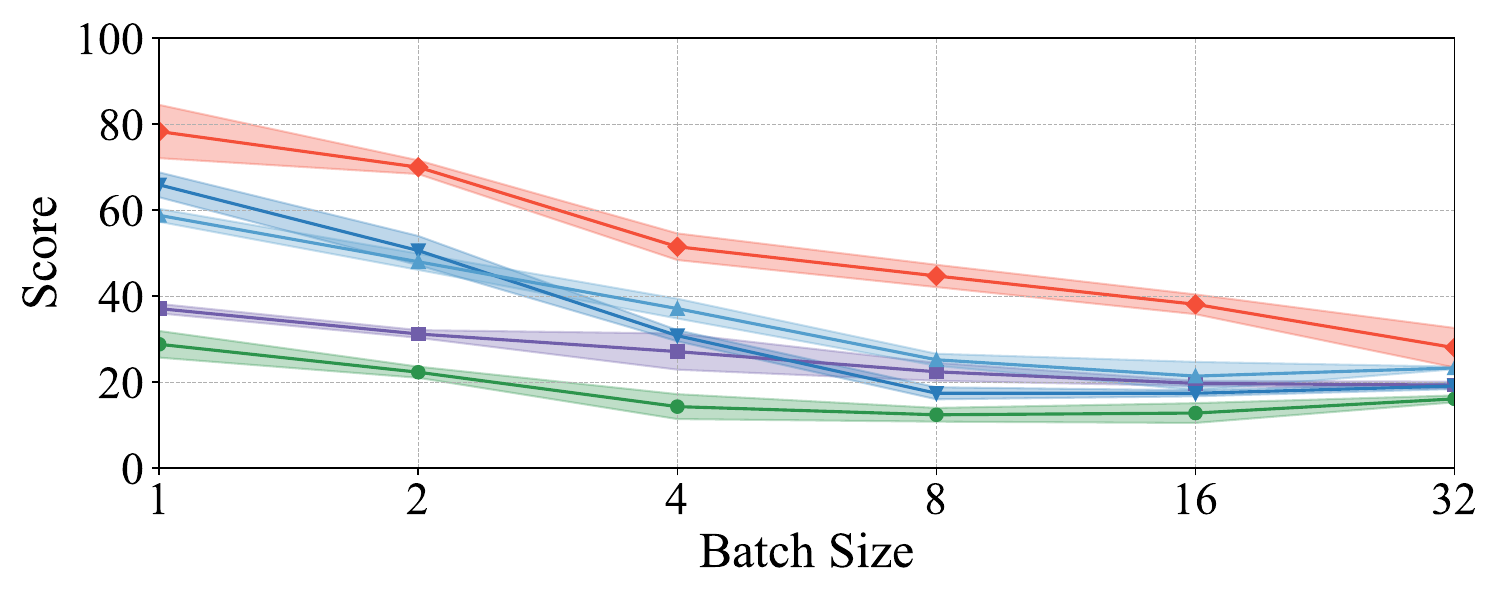}
    \end{subfigure}
    
    \begin{subfigure}[b]{0.017\linewidth}
        \raisebox{1.8\height}{\includegraphics[width=\linewidth]{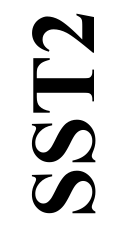}}
    \end{subfigure}
    \hfill
    \begin{subfigure}[b]{0.32\textwidth}
        \includegraphics[width=\textwidth]{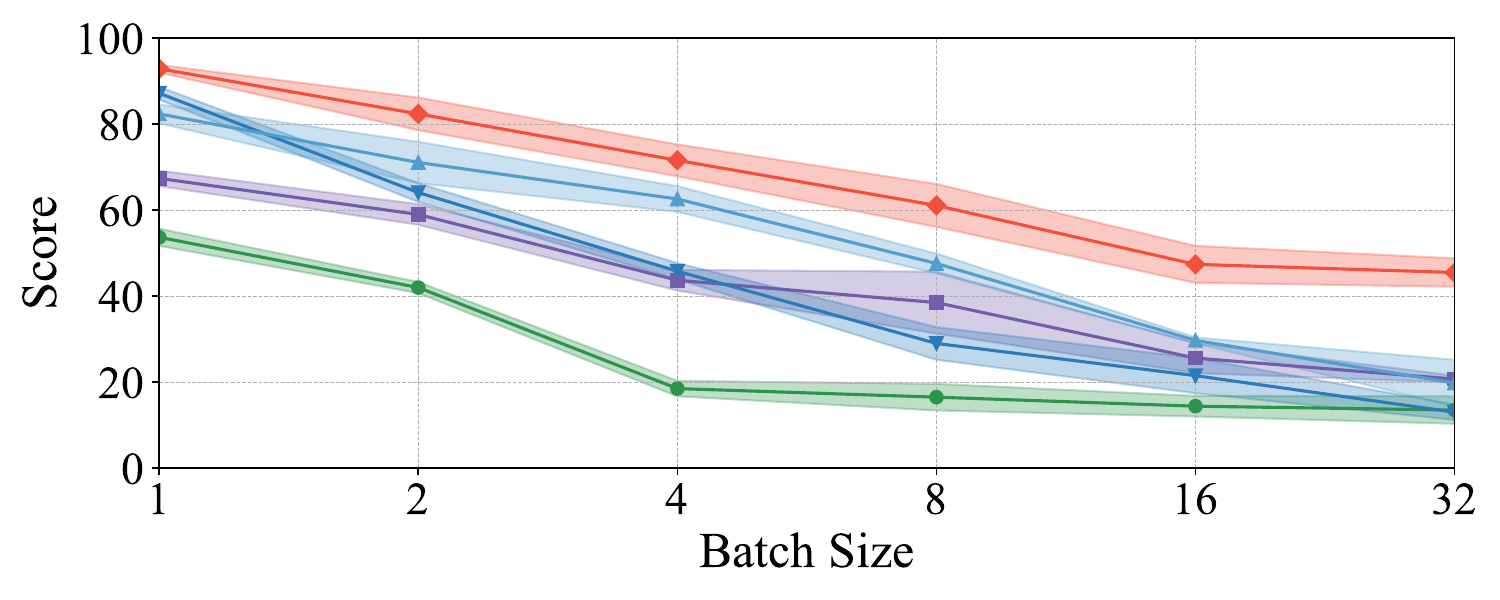}
    \end{subfigure}
    \hfill
    \begin{subfigure}[b]{0.32\textwidth}
        \includegraphics[width=\textwidth]{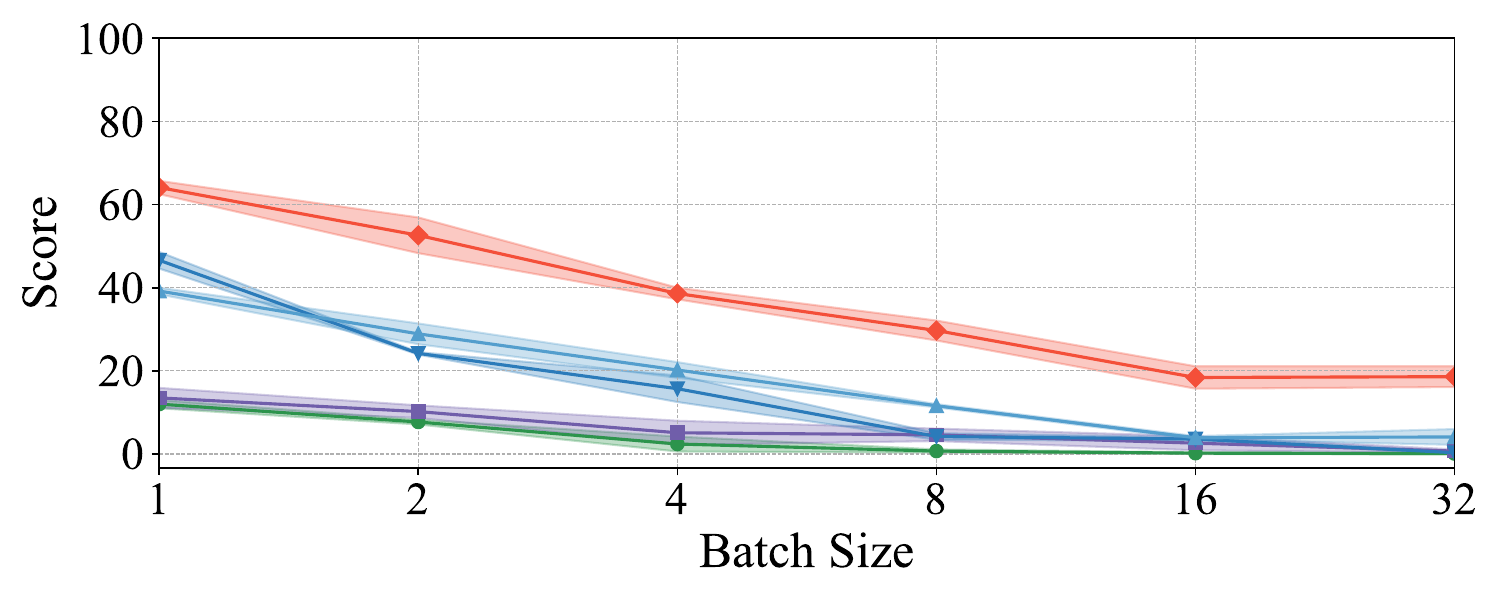}
    \end{subfigure}
    \hfill
    \begin{subfigure}[b]{0.32\textwidth}
        \includegraphics[width=\textwidth]{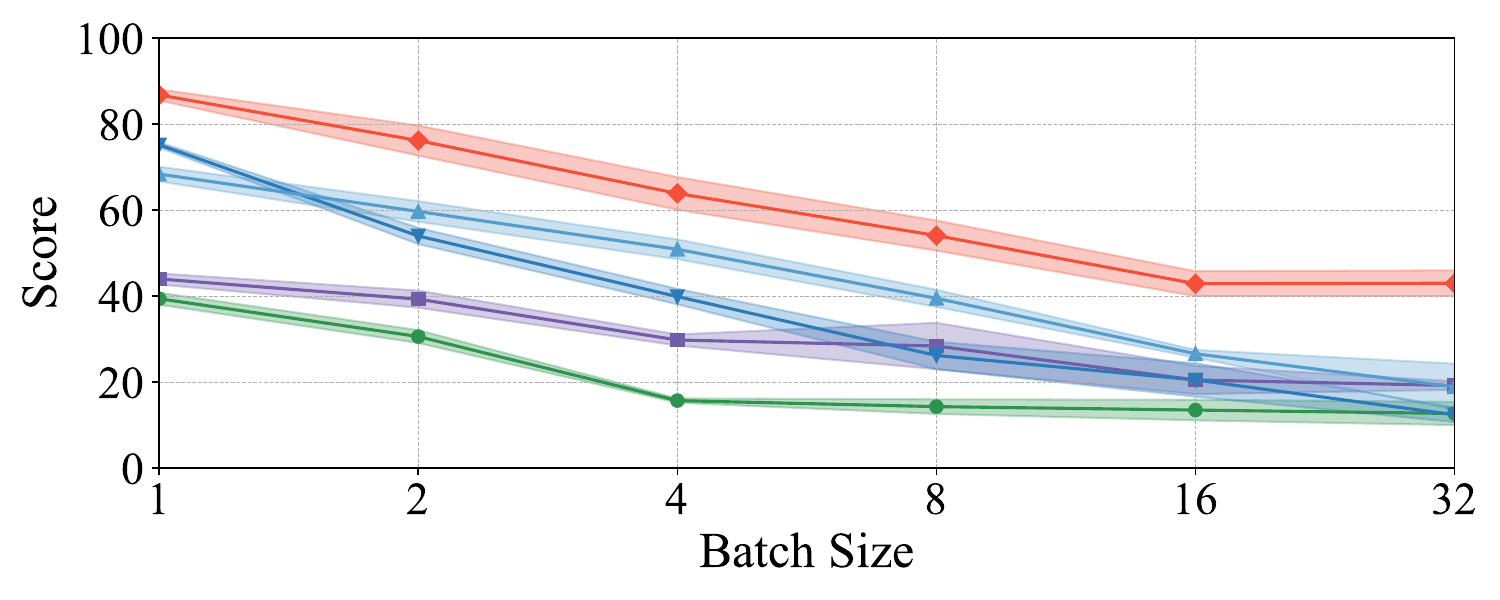}
    \end{subfigure}
    
    \begin{subfigure}[b]{0.017\linewidth}
        \raisebox{0.6\height}{
        \includegraphics[width=\linewidth]{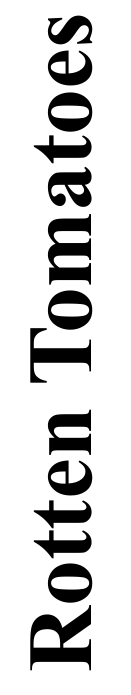}}
    \end{subfigure}
    \hfill
    \begin{subfigure}[b]{0.32\textwidth}
    \includegraphics[width=\textwidth]{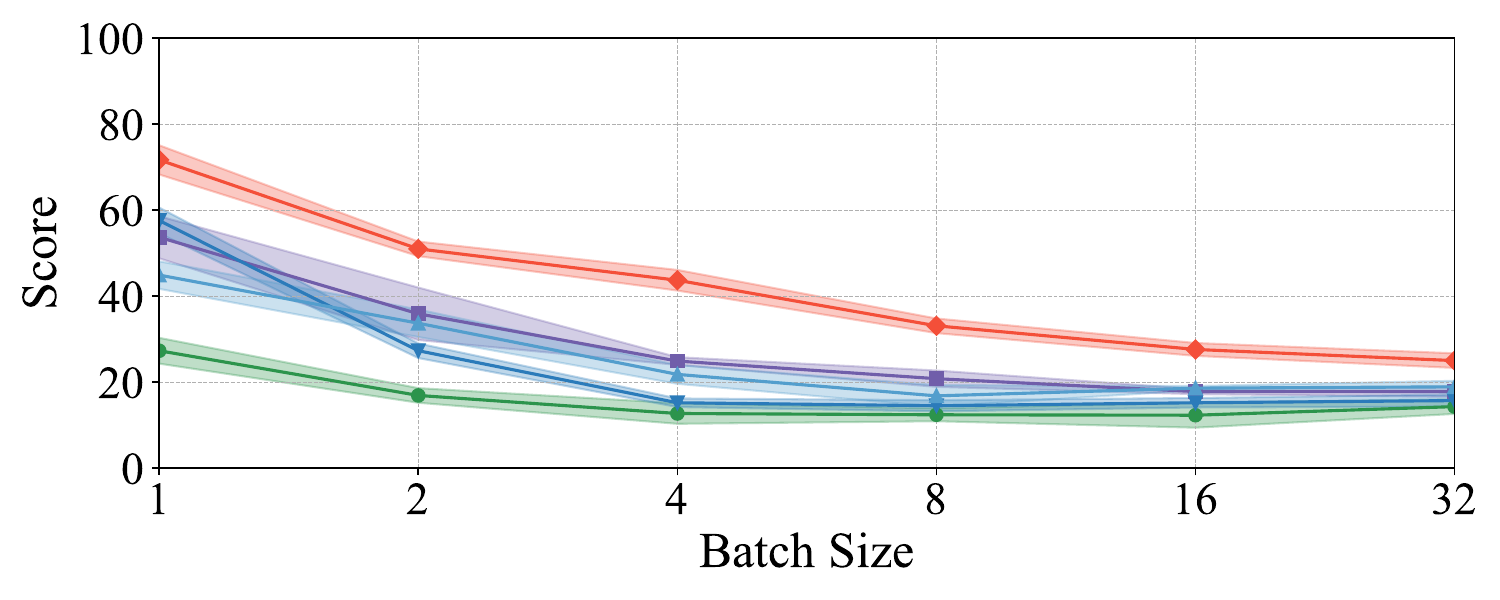}
        \caption{\; ROUGE-1}
    \end{subfigure}
    \hfill
    \begin{subfigure}[b]{0.32\textwidth}
        
        \includegraphics[width=\textwidth]{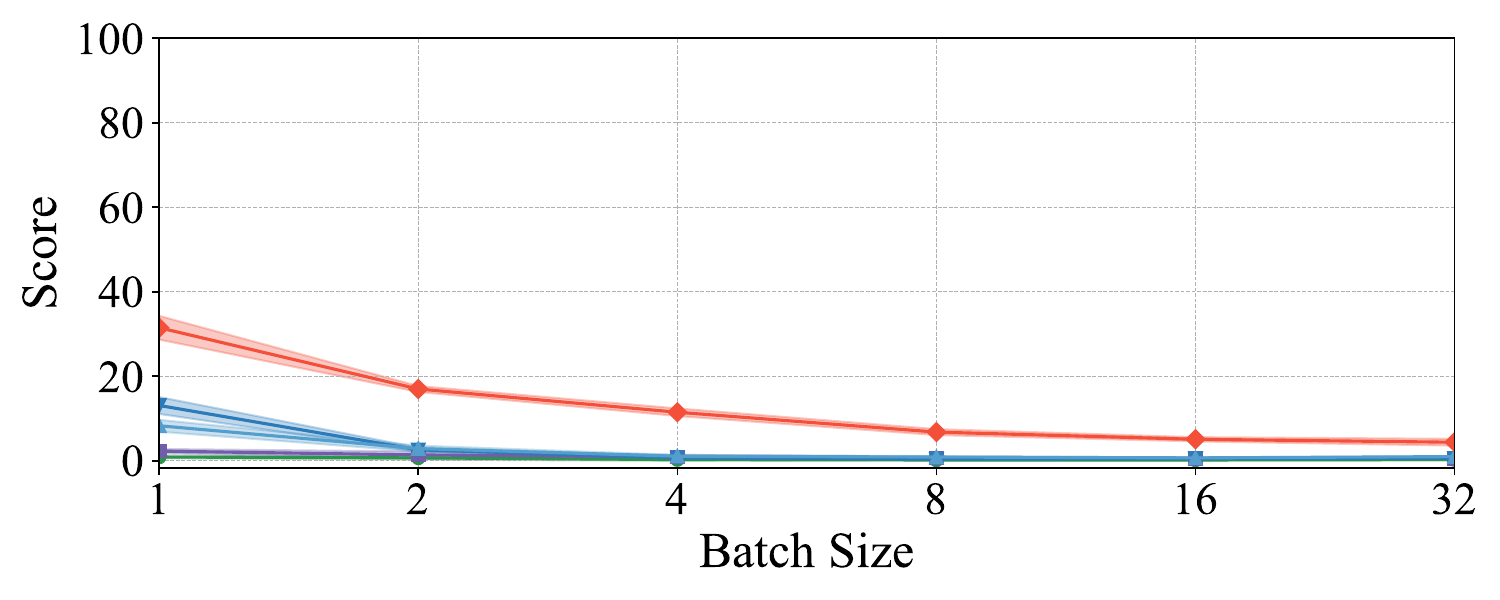}
        \caption{\; ROUGE-2}
    \end{subfigure}
    \hfill
    \begin{subfigure}[b]{0.32\textwidth}
        
        \includegraphics[width=\textwidth]{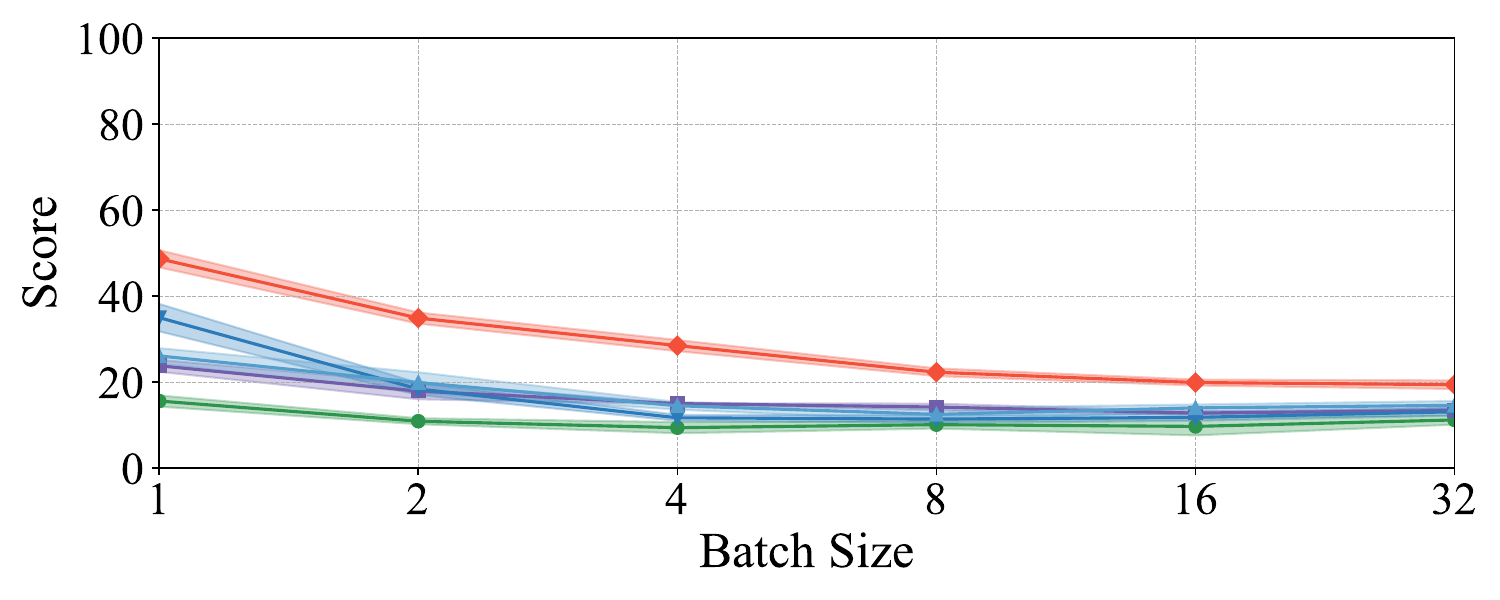}
        \caption{\; ROUGE-L}
    \end{subfigure}
    \hfill
    \begin{subfigure}[b]{1\textwidth}
    \centering
        \includegraphics[width=0.8\textwidth]{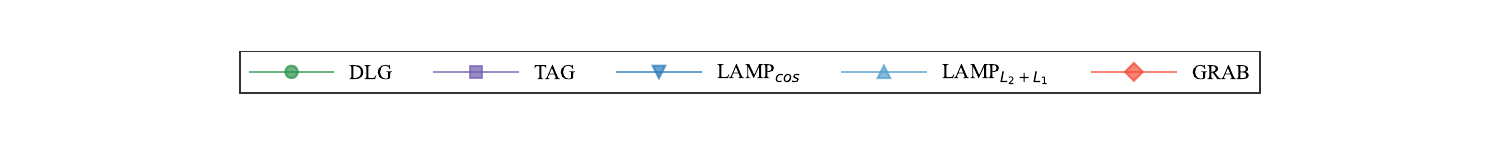}
    \end{subfigure}
    \caption{Performance in benchmark settings on the $\text{BERT}_{base}$ model.}
    \label{fig:benchmark_setting_performance}
\end{figure*}

%% file: figures/versatile_performance.tex
\begin{figure*}[h!]
    \centering
    
    \begin{subfigure}[b]{0.017\linewidth}
         \raisebox{1.5\height}{
        \includegraphics[width=\linewidth]{figures/new_result_figures/baselines/cola.pdf}}
    \end{subfigure}
    \hfill
    \begin{subfigure}[b]{0.32\textwidth}
        \includegraphics[width=\textwidth]{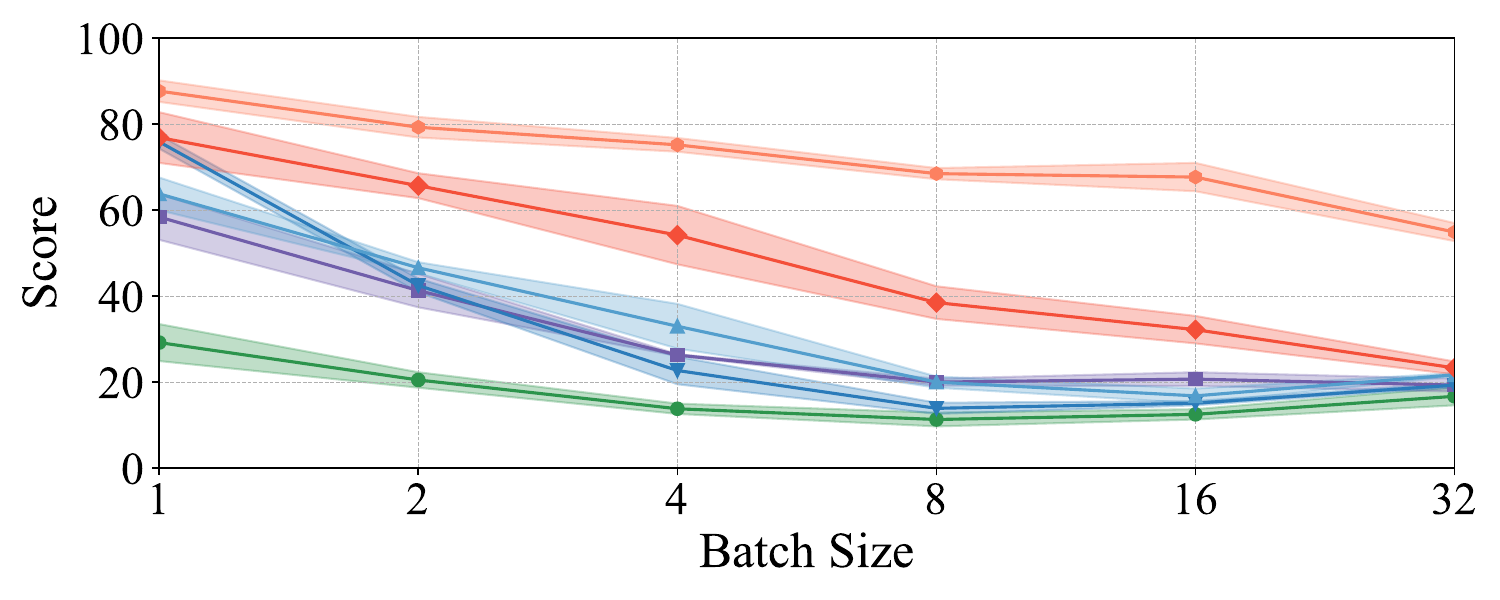}
    \end{subfigure}
    \hfill
    \begin{subfigure}[b]{0.32\textwidth}
        \includegraphics[width=\textwidth]{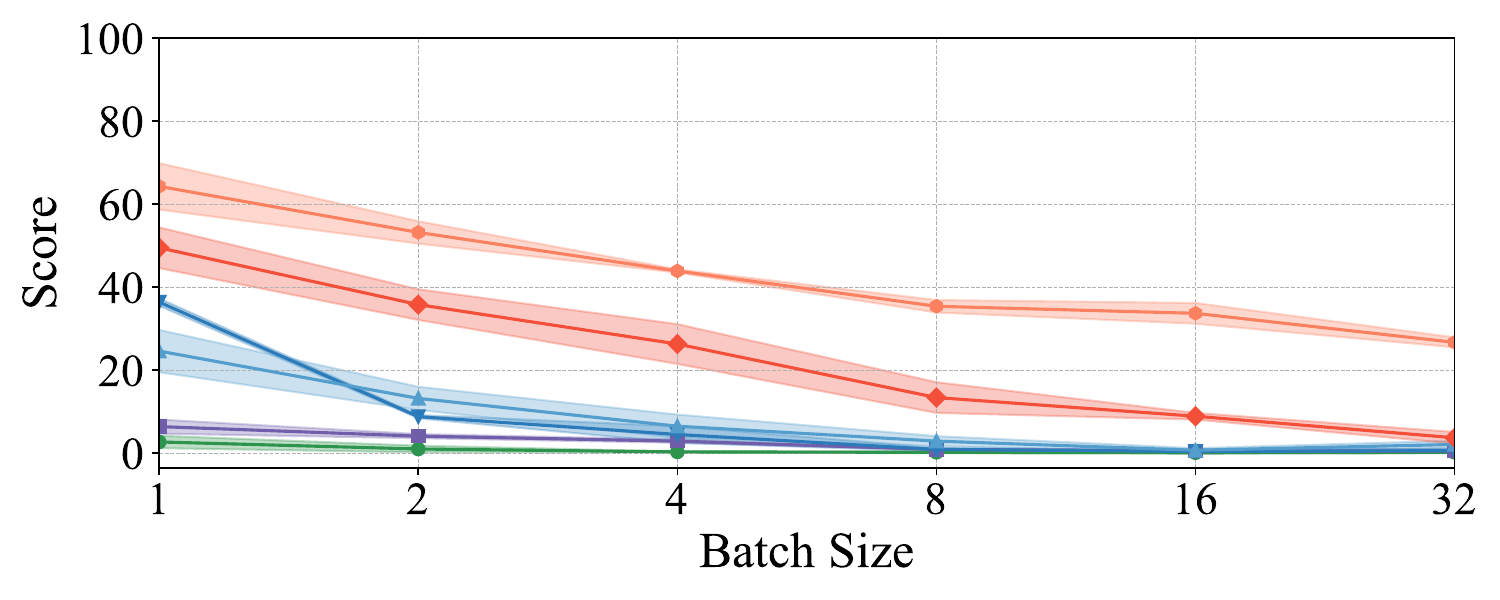}
    \end{subfigure}
    \hfill
    \begin{subfigure}[b]{0.32\textwidth}
        \includegraphics[width=\textwidth]{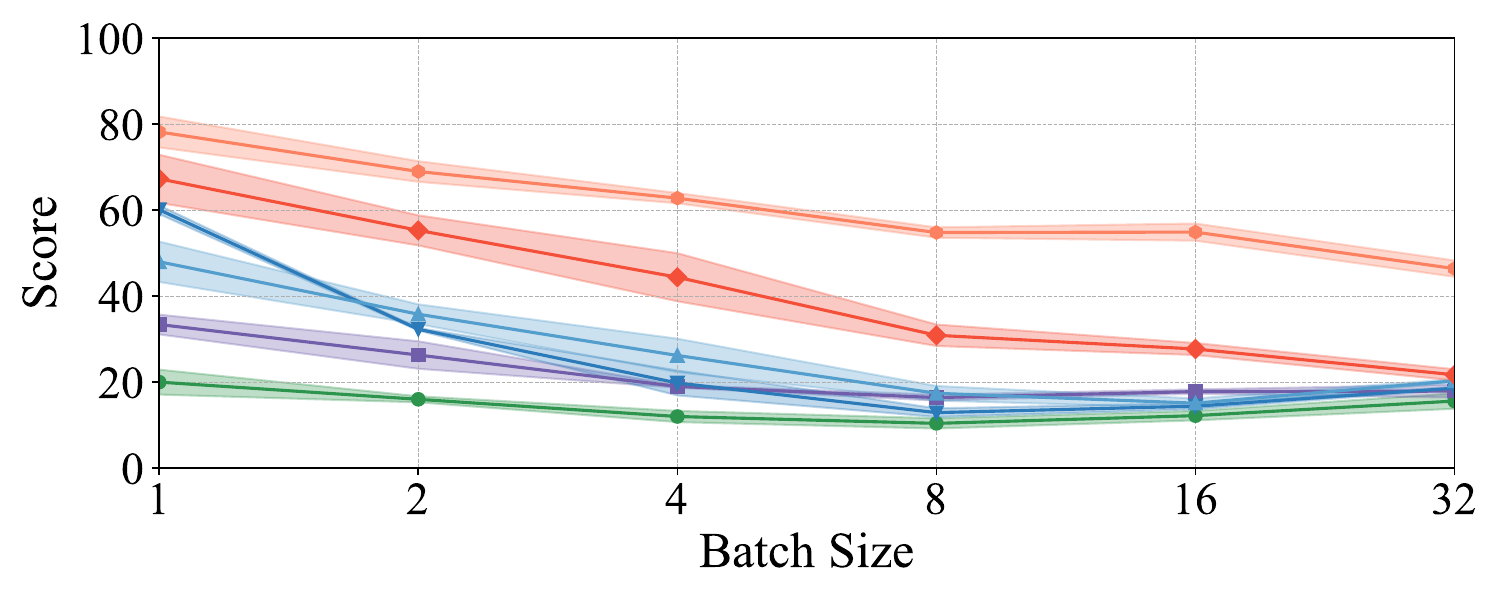}
    \end{subfigure}
    
    \begin{subfigure}[b]{0.017\linewidth}
        \raisebox{1.8\height}{
        \includegraphics[width=\linewidth]{figures/new_result_figures/baselines/sst2.pdf}}
    \end{subfigure}
    \hfill
    \begin{subfigure}[b]{0.32\textwidth}
        \includegraphics[width=\textwidth]{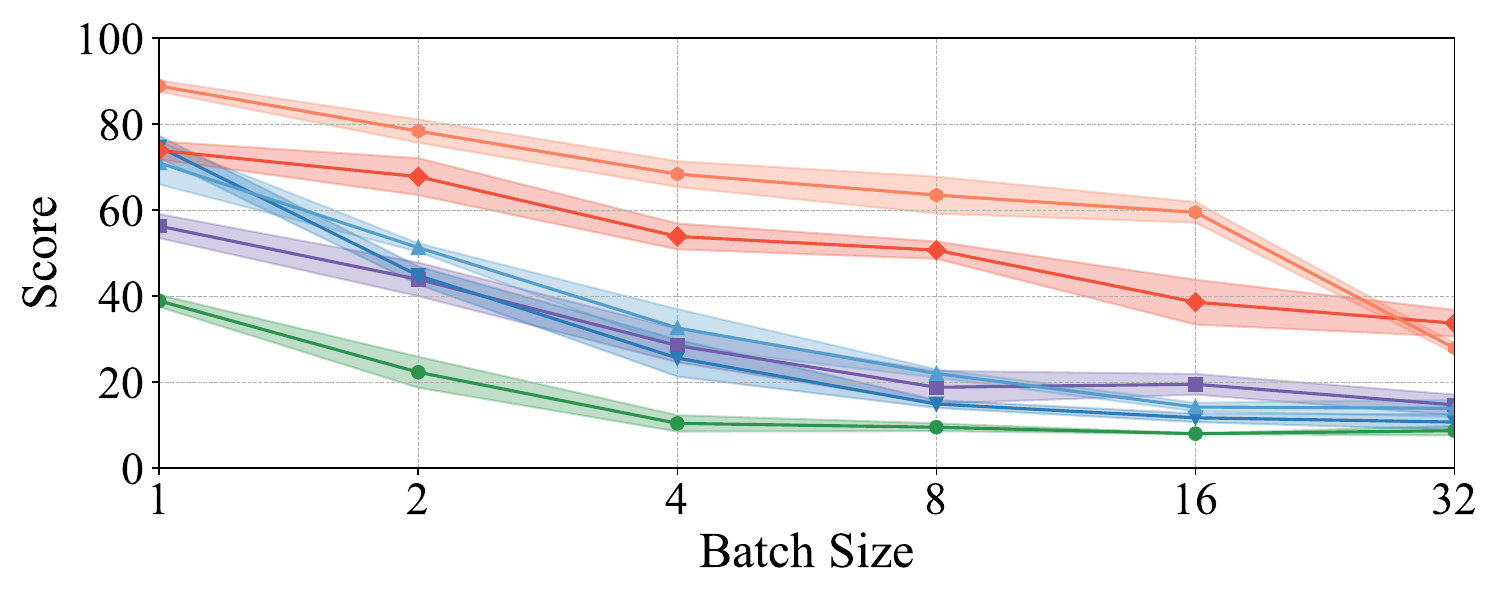}
    \end{subfigure}
    \hfill
    \begin{subfigure}[b]{0.32\textwidth}
        \includegraphics[width=\textwidth]{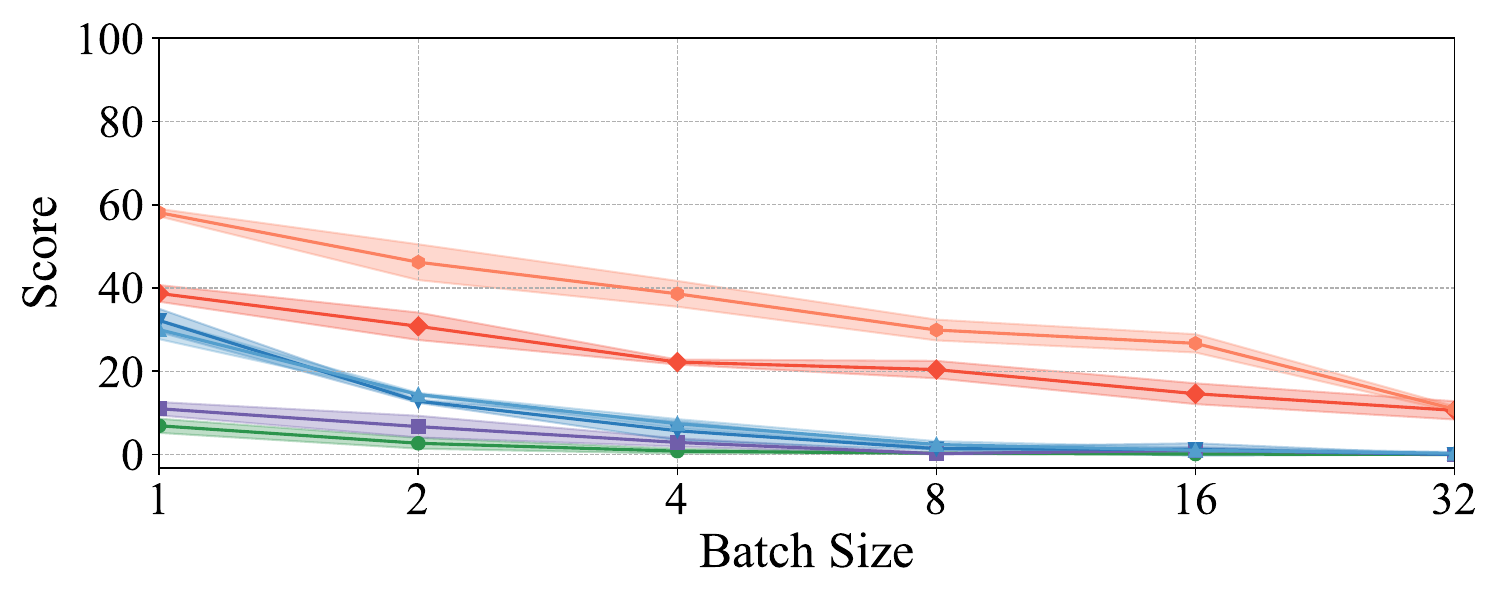}
    \end{subfigure}
    \hfill
    \begin{subfigure}[b]{0.32\textwidth}
        \includegraphics[width=\textwidth]{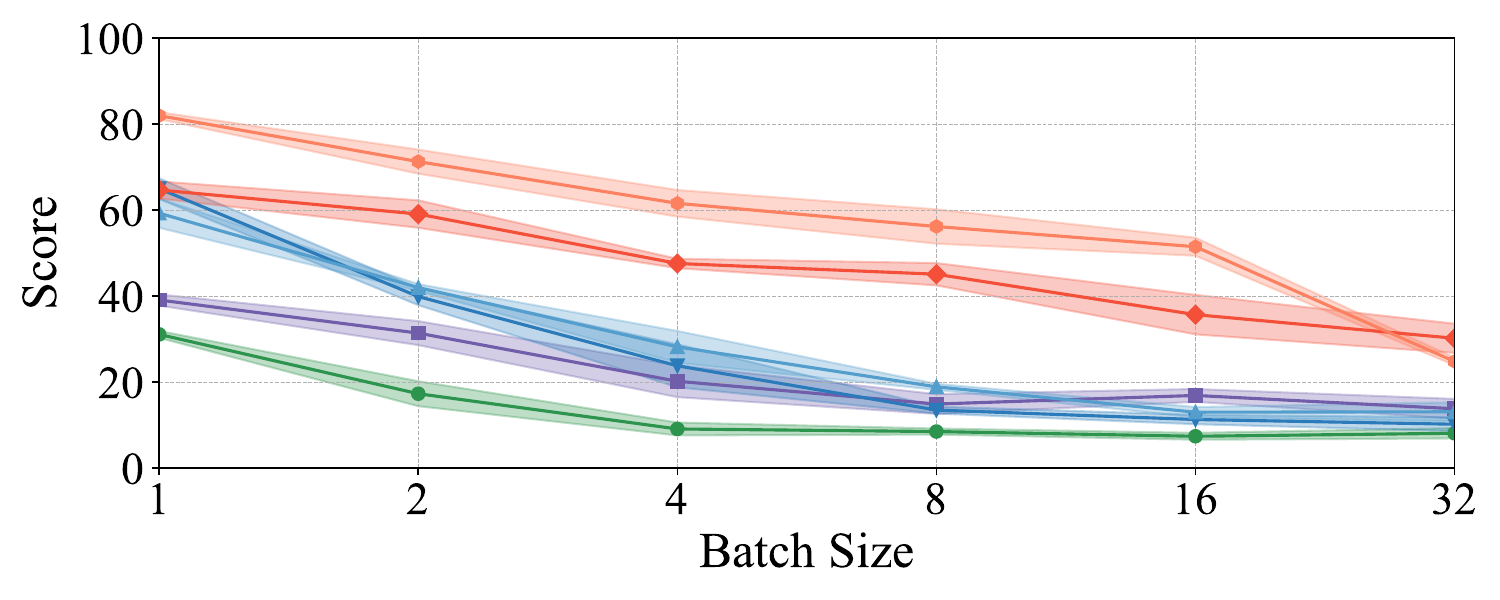}
    \end{subfigure}
    
    \begin{subfigure}[b]{0.017\linewidth}
        \raisebox{0.6\height}{
        \includegraphics[width=\linewidth]{figures/new_result_figures/baselines/RT.pdf}}
    \end{subfigure}
    \hfill
    \begin{subfigure}[b]{0.32\textwidth}
        
        \includegraphics[width=\textwidth]{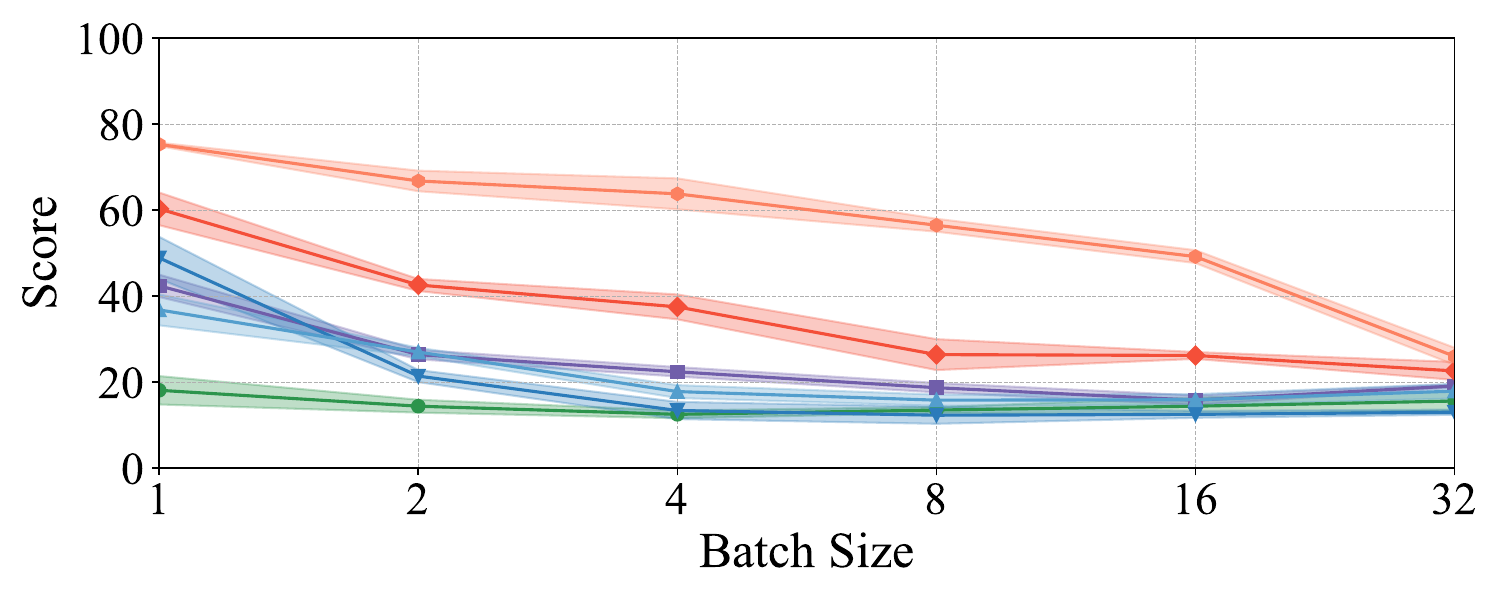}
        \caption{\; ROUGE-1}
    \end{subfigure}
    \hfill
    \begin{subfigure}[b]{0.32\textwidth}
        
        \includegraphics[width=\textwidth]{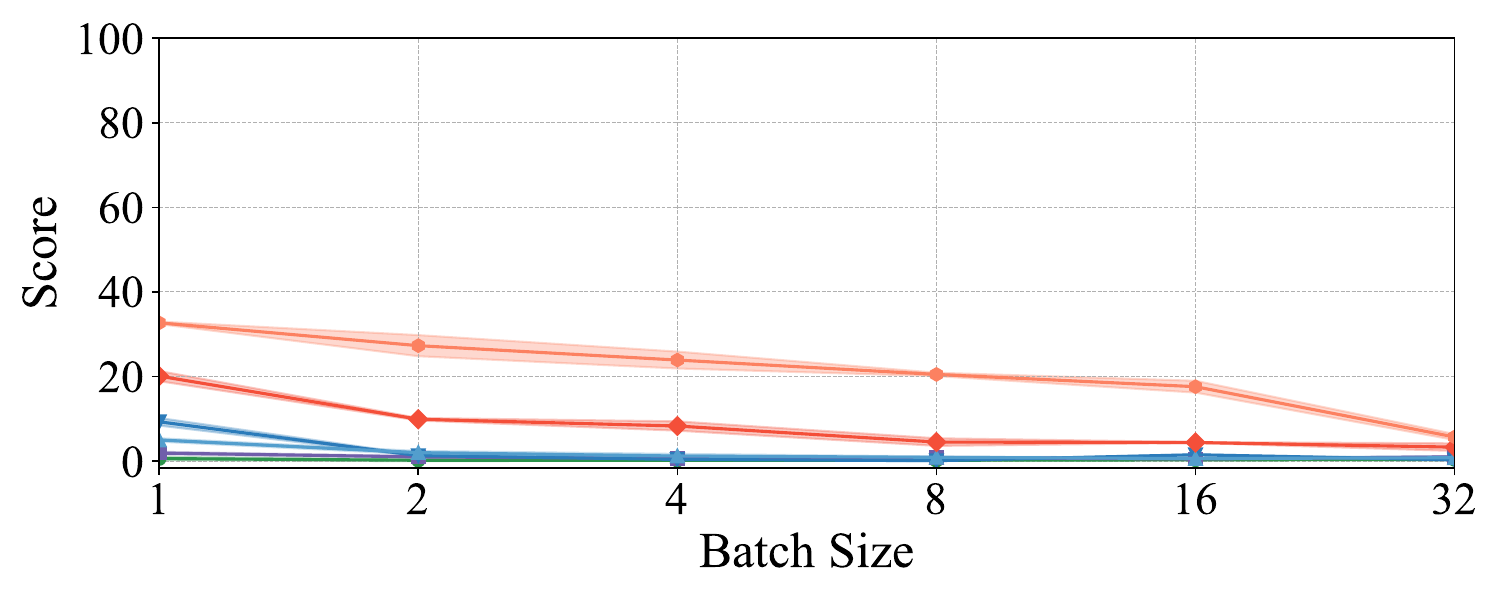}
        \caption{\; ROUGE-2}
    \end{subfigure}
    \hfill
    \begin{subfigure}[b]{0.32\textwidth}
        
        \includegraphics[width=\textwidth]{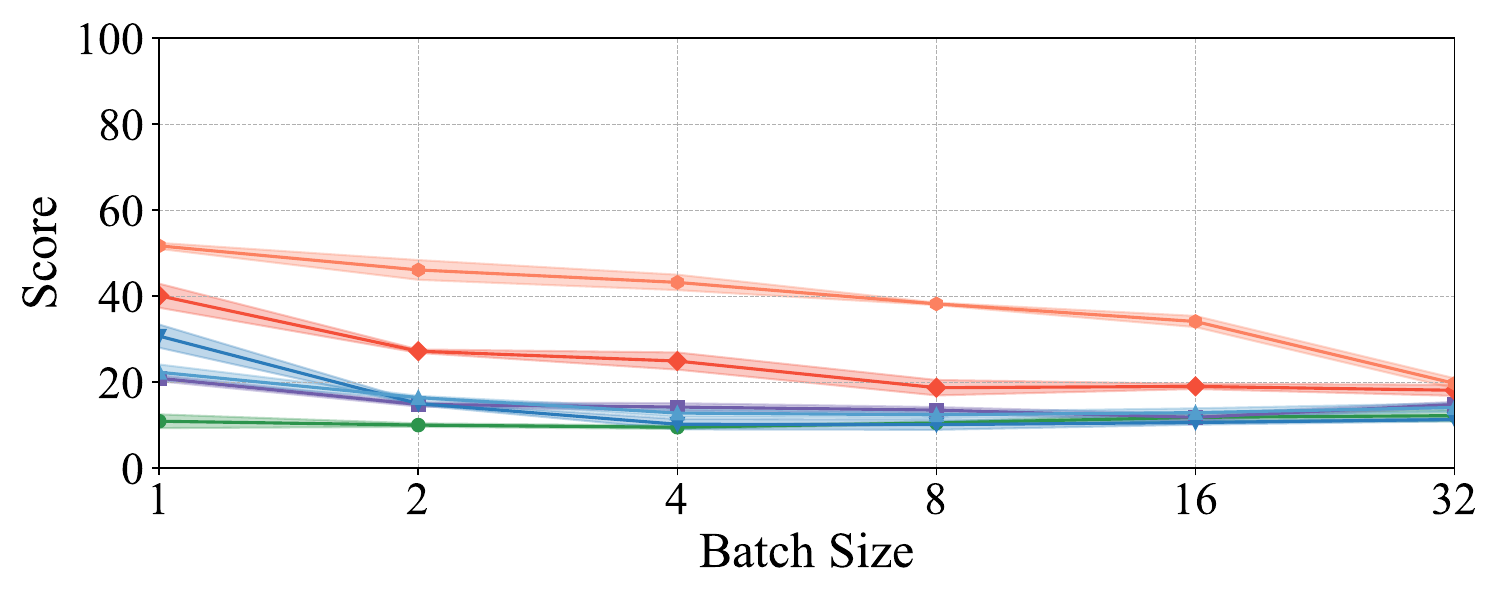}
        \caption{\; ROUGE-L}
    \end{subfigure}
    \hfill
    \begin{subfigure}[b]{1\textwidth}
    \centering
        \includegraphics[width=0.8\textwidth]{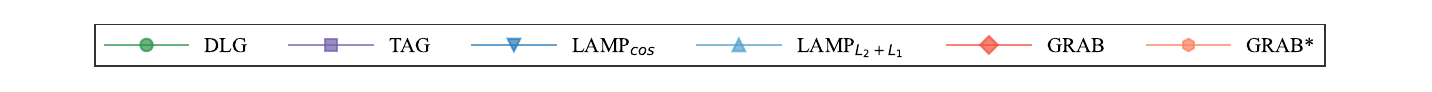}
    \end{subfigure}
    \caption{Performance in practical settings on the $\text{BERT}_{base}$ model. $\text{\codename}^{*}$ is the variant with dropout mask learning.}
    \label{fig:versatile_setting_performance}
\end{figure*}

%% file: tables/RQ3_1.tex
\begin{table*}[!t]
\centering
\newcommand{\threecol}[1]{\multicolumn{3}{c}{#1}}
\newcommand{\skiplen}{0.008\linewidth} 
\renewcommand{\arraystretch}{0.9}

\caption{Performance in practical settings on various BERT and RoBERTa model sizes.} 
\label{table:RQ3_1_1}

\begingroup 
\setlength{\tabcolsep}{6pt} %

\begin{threeparttable}
\scriptsize
\begin{tabular}{l rrr p{\skiplen} rrr p{\skiplen} rrr p{\skiplen} rrr p{\skiplen} rrr p{\skiplen} rrr} 
\toprule
& \threecol{$\text{BERT}_{tiny}$}&& \threecol{$\text{BERT}_{base}$}&& \threecol{$\text{BERT}_{large}$}\\
\cmidrule(l{5pt}r{5pt}){2-4} 
\cmidrule(l{5pt}r{5pt}){6-8} 
\cmidrule(l{5pt}r{1pt}){10-12}
& \makecell[c]{R-1} & \makecell[c]{R-2} & \makecell[c]{R-L} && 
\makecell[c]{R-1} & \makecell[c]{R-2} & \makecell[c]{R-L} && 
\makecell[c]{R-1} & \makecell[c]{R-2} & \makecell[c]{R-L} \\ \midrule
DLG & 
3.9~$\pm$~0.5 & 0.0~$\pm$~0.0 & 3.6~$\pm$~0.5 && 
29.2~$\pm$~4.3 & 2.7~$\pm$~1.5 & 20.0~$\pm$~2.9 && 
5.6~$\pm$~1.3 & 0.8~$\pm$~0.6 & 4.9~$\pm$~0.9 \\
TAG & 
14.2~$\pm$~3.1 & 0.9~$\pm$~0.9 & 12.1~$\pm$~2.6 && 
58.4~$\pm$~5.3 & 6.4~$\pm$~1.7 & 33.4~$\pm$~2.3 && 
11.9~$\pm$~0.4 & 1.0~$\pm$~0.6 & 8.4~$\pm$~0.7 \\
LAMP$_{cos}$ & 
84.1~$\pm$~0.5 & 42.9~$\pm$~1.1 & 66.1~$\pm$~0.9 && 
75.9~$\pm$~1.6 & 36.4~$\pm$~0.9 & 60.1~$\pm$~1.0 && 
17.3~$\pm$~1.3 & 4.2~$\pm$~0.9 & 15.2~$\pm$~0.5 \\
LAMP$_{L_{2}+L_{1}}$ & 
79.8~$\pm$~1.2 & 34.2~$\pm$~2.3 & 60.2~$\pm$~2.3 && 
63.8~$\pm$~3.8 & 24.6~$\pm$~5.1 & 48.0~$\pm$~4.7 && 
1.3~$\pm$~0.7 & 0.1~$\pm$~0.1 & 1.3~$\pm$~0.7 \\
\codename & 
88.3~$\pm$~2.0 & 60.3~$\pm$~4.5 & 79.8~$\pm$~2.1 && 
76.9~$\pm$~5.9 & 49.5~$\pm$~4.9 & 67.3~$\pm$~5.6 && 
14.3~$\pm$~6.1 & 4.4~$\pm$~3.1 & 12.6~$\pm$~5.1 \\
\codename$^*$ & 
\textbf{90.2}~$\pm$~1.5 & \textbf{60.6}~$\pm$~0.5 & \textbf{80.0}~$\pm$~1.3 && 
\textbf{87.7}~$\pm$~2.5 & \textbf{64.3}~$\pm$~5.6 & \textbf{78.2}~$\pm$~3.6 && 
\textbf{70.2}~$\pm$~4.3 & \textbf{32.2}~$\pm$~4.3 & \textbf{55.0}~$\pm$~2.8 \\
\midrule
& \threecol{$\text{RoBERTa}_{tiny}$}&& \threecol{$\text{RoBERTa}_{base}$}&& \threecol{$\text{RoBERTa}_{large}$}\\
\cmidrule(l{5pt}r{5pt}){2-4} 
\cmidrule(l{5pt}r{5pt}){6-8} 
\cmidrule(l{5pt}r{1pt}){10-12}
\codename & 
71.2~$\pm$~1.1 & 37.1~$\pm$~1.1 & 62.7~$\pm$~0.7 && 
43.6~$\pm$~3.2 & 22.5~$\pm$~2.3 & 38.7~$\pm$~3.5 && 
1.1~$\pm$~0.1 & 0.0~$\pm$~0.0 & 1.1~$\pm$~0.1 \\
\codename$^*$ & 
\textbf{82.8}~$\pm$~1.5 & \textbf{58.8}~$\pm$~3.0 & \textbf{77.2}~$\pm$~1.6 && 
\textbf{71.6}~$\pm$~6.3 & \textbf{36.8}~$\pm$~8.5 & \textbf{60.1}~$\pm$~5.9 && 
\textbf{31.7}~$\pm$~8.1 & \textbf{11.0}~$\pm$~4.5 & \textbf{26.9}~$\pm$~7.3 \\
\bottomrule
\end{tabular}
\begin{tablenotes}
    \scriptsize
    \item[$*$] with dropout mask learning.
\end{tablenotes}
\end{threeparttable}

\endgroup
\vspace{-6pt}
\end{table*}






\begin{table}[!t]
\centering
\renewcommand{\arraystretch}{0.9}

\caption{Performance of tuned hyperparameters.}
\label{table:RQ3_1_3}
\scriptsize
\resizebox{1\linewidth}{!}{

\setlength{\tabcolsep}{3pt}
\begin{threeparttable}
\begin{tabular}{lrrrrrr}
\toprule
& \multicolumn{3}{c}{$\text{BERT}_{large}$} & \multicolumn{3}{c}{$\text{RoBERTa}_{large}$} \\
\cmidrule(l{5pt}r{5pt}){2-4} \cmidrule(l{5pt}r{5pt}){5-7}
& \makecell[c]{R-1} & \makecell[c]{R-2} & \makecell[c]{R-L} & \makecell[c]{R-1} & \makecell[c]{R-2} & \makecell[c]{R-L} \\
\midrule
DLG & 51.4~$\pm$~5.0 & 7.5~$\pm$~2.4 & 33.1~$\pm$~2.8 & - & - & - \\
TAG & 61.1~$\pm$~2.6 & 8.0~$\pm$~1.6 & 37.6~$\pm$~2.1 & - & - & - \\
LAMP$_{cos}$ & 76.4~$\pm$~2.1 & 34.3~$\pm$~3.0 & 60.2~$\pm$~2.2 & - & - & - \\
LAMP$_{L_{2}+L_{1}}$ & 70.9~$\pm$~3.4 & 25.1~$\pm$~3.4 & 52.9~$\pm$~2.6 & - & - & - \\
\codename & 64.6~$\pm$~2.5 & 29.3~$\pm$~1.7 & 52.5~$\pm$~2.3 & 7.1~$\pm$~2.1 & 1.0~$\pm$~0.8 & 6.0~$\pm$~1.4 \\
\codename$^*$ & \textbf{84.8}~$\pm$~1.8 & \textbf{50.2}~$\pm$~4.9 & \textbf{69.3}~$\pm$~3.5 & \textbf{74.1}~$\pm$~1.0 & \textbf{33.2}~$\pm$~3.2 & \textbf{59.2}~$\pm$~0.9 \\
\bottomrule
\end{tabular}
\begin{tablenotes}
    \scriptsize
    \item[$*$] with dropout mask learning.
\end{tablenotes}
\end{threeparttable}
}
\vspace{-15pt}
\end{table}

%% file: figures/dropout_performance.tex
\begin{figure*}[t!]
    \centering
    
    \begin{subfigure}[b]{0.017\linewidth}
          \raisebox{2.3\height}{\includegraphics[width=\linewidth]{figures/new_result_figures/baselines/cola.pdf}}
    \end{subfigure}
    \hfill
    \begin{subfigure}[b]{0.32\textwidth}
        \includegraphics[width=\textwidth]{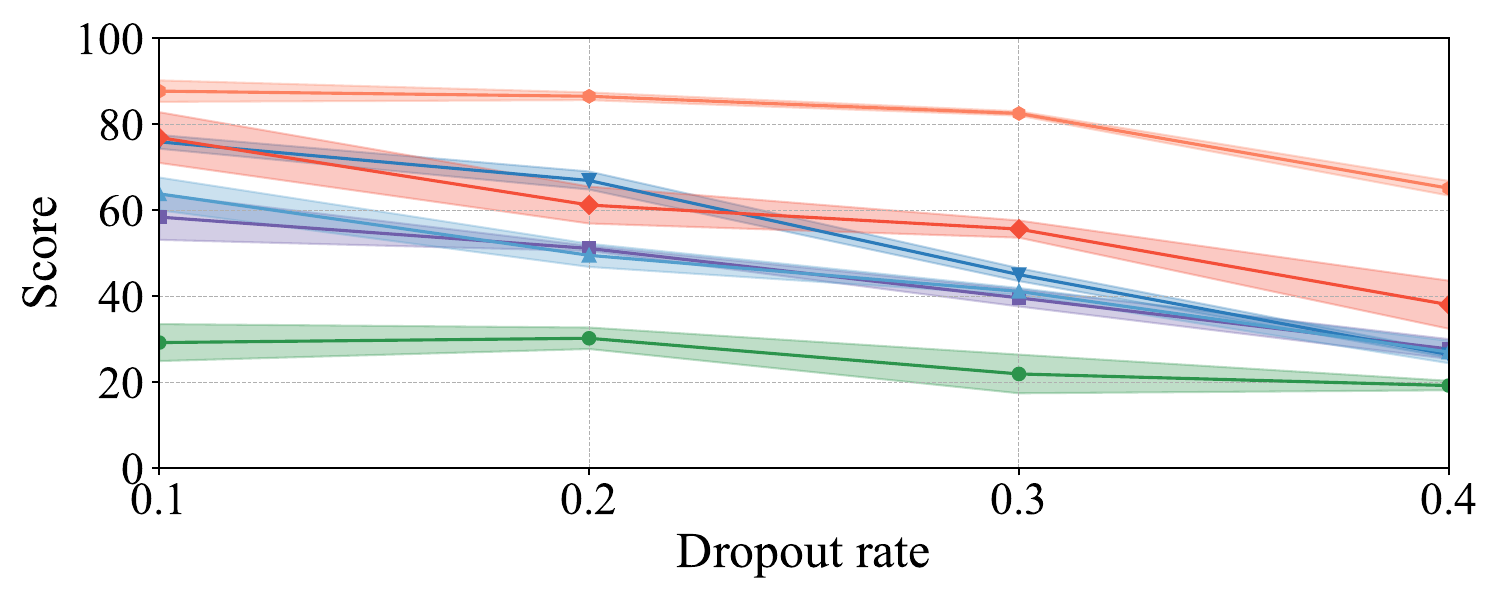}
        \caption{\; ROUGE-1}
    \end{subfigure}
    \hfill
    \begin{subfigure}[b]{0.32\textwidth}
        \includegraphics[width=\textwidth]{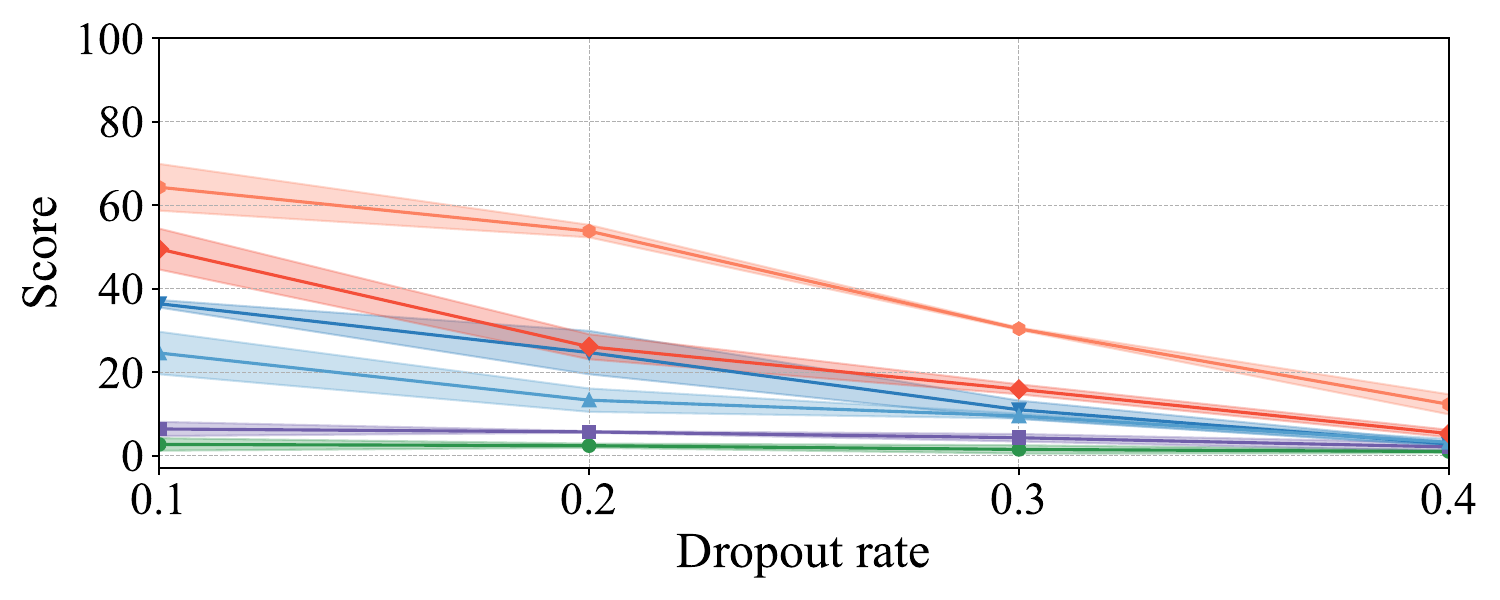}
        \caption{\; ROUGE-2}
    \end{subfigure}
    \hfill
    \begin{subfigure}[b]{0.32\textwidth}
        \includegraphics[width=\textwidth]{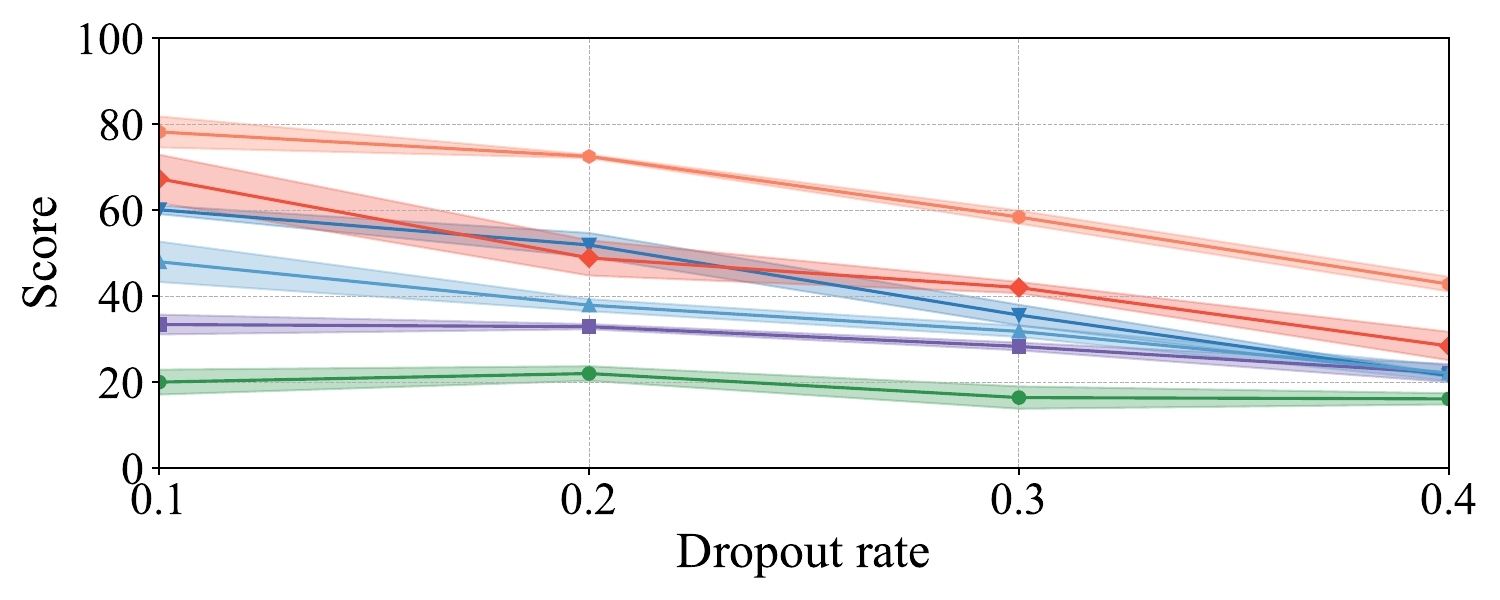}
        \caption{\; ROUGE-L}
    \end{subfigure}
    
    \hfill
    \begin{subfigure}[b]{1\textwidth}
    \centering
        \includegraphics[width=0.8\textwidth]{figures/new_result_figures/versatile/legend.pdf}
    \end{subfigure}
    
    \caption{Performance in practical settings with different dropout rates. $\text{\codename}^{*}$ is the variant with dropout mask learning.}
    \label{fig:dropout_performance}
\end{figure*}

%% file: tables/RQ3_5_7.tex
\begin{table*}[!t]
\centering
\scriptsize
\newcommand{\threecol}[1]{\multicolumn{3}{c}{#1}} 
\newcommand{\skiplen}{0.008\linewidth} 
\renewcommand{\arraystretch}{1.}
\caption{Ablation study on assumptions of known labels and known exact sequence lengths.}
\label{table:RQ3_5_7}

\begingroup
\setlength{\tabcolsep}{6pt}

\begin{threeparttable}
\begin{tabular}{l rrr p{\skiplen} rrr p{\skiplen} rrr p{\skiplen} rrr p{\skiplen} rrr}
\toprule
& \threecol{B = 1 } && \threecol{B = 2 } && \threecol{B = 4 }\\
\cmidrule(l{5pt}r{5pt}){2-4} 
\cmidrule(l{5pt}r{5pt}){6-8} 
\cmidrule(l{5pt}r{1pt}){10-12}
& \makecell[c]{R-1} & \makecell[c]{R-2} & \makecell[c]{R-L} && 
\makecell[c]{R-1} & \makecell[c]{R-2} & \makecell[c]{R-L} && 
\makecell[c]{R-1} & \makecell[c]{R-2} & \makecell[c]{R-L} \\ 
\midrule
\codename$^1$ 
& - & - & - && 
80.7~$\pm$~3.1 & 53.4~$\pm$~2.3 & 68.9~$\pm$~1.8 && 
76.5~$\pm$~0.6 & 47.8~$\pm$~1.0 & 63.9~$\pm$~1.0 \\
\codename$^2$ & 
\textbf{88.6}~$\pm$~1.6 & \textbf{65.7}~$\pm$~1.4 & \textbf{78.8}~$\pm$~1.1 && 
78.8~$\pm$~0.9 & 51.9~$\pm$~2.2 & 67.4~$\pm$~0.4 && 
67.7~$\pm$~2.3 & 36.7~$\pm$~2.8 & 55.5~$\pm$~1.7 \\
\codename$^3$ 
& - & - & - && 
\textbf{84.9}~$\pm$~1.2 & \textbf{56.8}~$\pm$~1.2 & \textbf{72.3}~$\pm$~1.2 && 
\textbf{79.2}~$\pm$~0.7 & \textbf{52.8}~$\pm$~3.0 & \textbf{66.9}~$\pm$~2.4 \\
\codename$^4$ & 
87.7~$\pm$~2.5 & 64.3~$\pm$~5.6 & 78.2~$\pm$~3.6 && 
79.3~$\pm$~2.4 & 53.2~$\pm$~2.7 & 69.0~$\pm$~2.4 && 
75.2~$\pm$~1.6 & 43.9~$\pm$~0.4 & 62.8~$\pm$~1.2 \\
\midrule
& \threecol{B = 8 } && \threecol{B = 16 } && \threecol{B = 32 }\\
\midrule
\codename$^1$ & 
71.0~$\pm$~0.5 & 43.6~$\pm$~1.3 & 58.8~$\pm$~0.6 && 
65.5~$\pm$~1.4 & \textbf{40.7}~$\pm$~2.1 & 53.8~$\pm$~1.7 && 
53.8~$\pm$~1.5 & 30.9~$\pm$~2.5 & 44.8~$\pm$~1.7 \\
\codename$^2$ & 
65.8~$\pm$~0.8 & 33.8~$\pm$~2.3 & 53.3~$\pm$~1.1 && 
62.4~$\pm$~1.6 & 32.1~$\pm$~1.1 & 51.4~$\pm$~0.9 && 
53.8~$\pm$~1.1 & 23.4~$\pm$~1.1 & 44.7~$\pm$~1.1 \\
\codename$^3$ & 
\textbf{73.6}~$\pm$~1.7 & \textbf{45.6}~$\pm$~1.7 & \textbf{58.9}~$\pm$~2.2 && 
62.8~$\pm$~2.7 & 38.0~$\pm$~4.1 & 52.2~$\pm$~2.9 && 
\textbf{55.9}~$\pm$~1.7 & \textbf{31.3}~$\pm$~3.1 & 45.9~$\pm$~2.1 \\
\codename$^4$ & 
68.5~$\pm$~1.3 & 35.4~$\pm$~1.5 & 54.8~$\pm$~1.2 && 
\textbf{67.7}~$\pm$~3.3 & 33.7~$\pm$~2.5 & \textbf{54.9}~$\pm$~2.0 && 
54.9~$\pm$~2.1 & 26.7~$\pm$~1.2 & \textbf{46.4}~$\pm$~1.9 \\
\bottomrule
\end{tabular}
\begin{tablenotes}
    \scriptsize
    \item
    $^1$
    Without the assumptions of known labels or sequence lengths.
    $^2$
    Without the assumption of known labels.
    $^3$
    Without the assumption of known sequence lengths.
    $^4$
    With both assumptions of known labels and known sequence lengths.
\end{tablenotes}
\end{threeparttable}

\endgroup
\vspace{-10pt}
\end{table*}

%% file: figures/noise_performance.tex
\begin{figure*}[h!]
    \centering
    
 \begin{subfigure}[b]{0.017\linewidth}
         \raisebox{2.3\height}{
        \includegraphics[width=\linewidth]{figures/new_result_figures/baselines/cola.pdf}}
    \end{subfigure}
    \hfill
    \begin{subfigure}[b]{0.32\textwidth}
        \includegraphics[width=\textwidth]{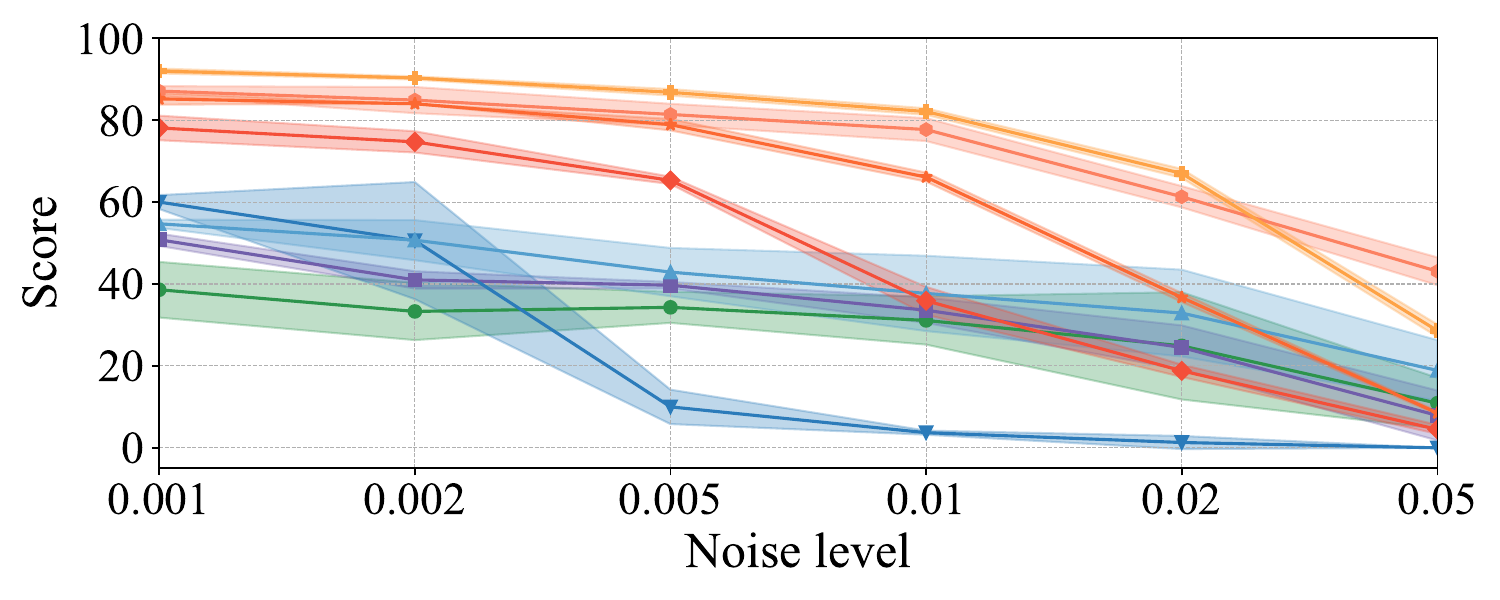}
        \caption{\; ROUGE-1}
    \end{subfigure}
    \hfill
    \begin{subfigure}[b]{0.32\textwidth}
        \includegraphics[width=\textwidth]{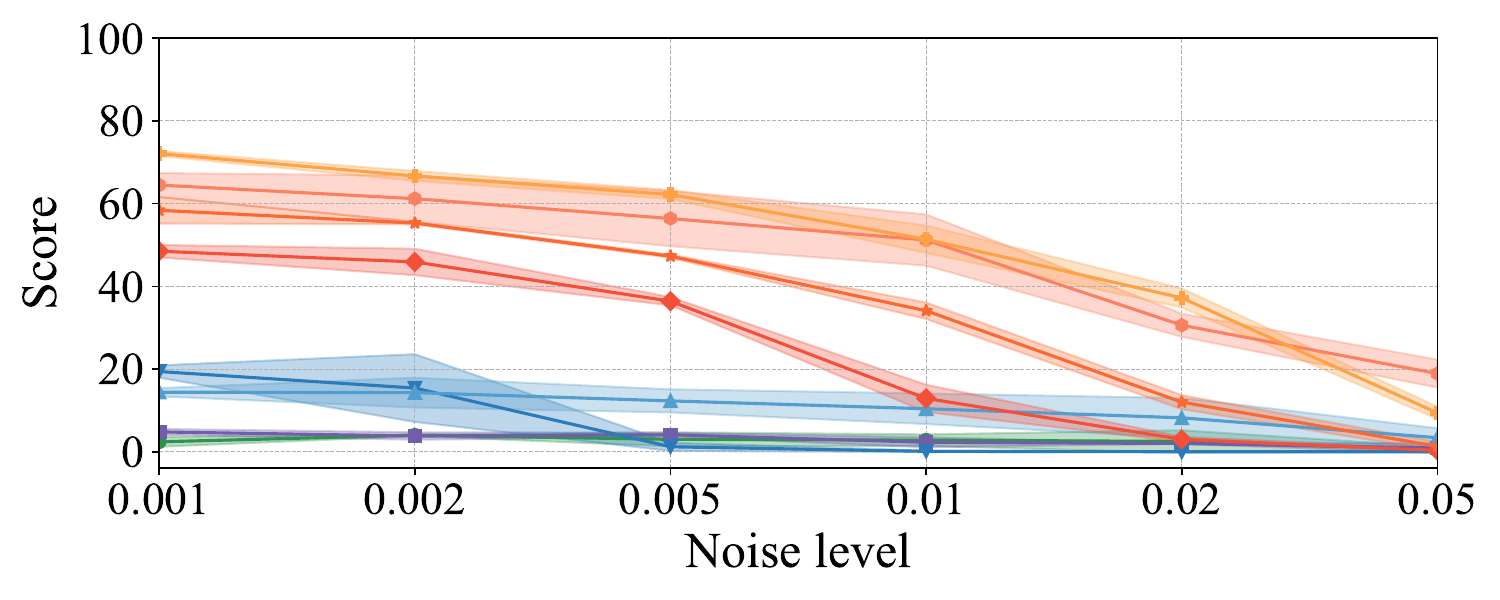}
        \caption{\; ROUGE-2}
    \end{subfigure}
    \hfill
    \begin{subfigure}[b]{0.32\textwidth}
        \includegraphics[width=\textwidth]{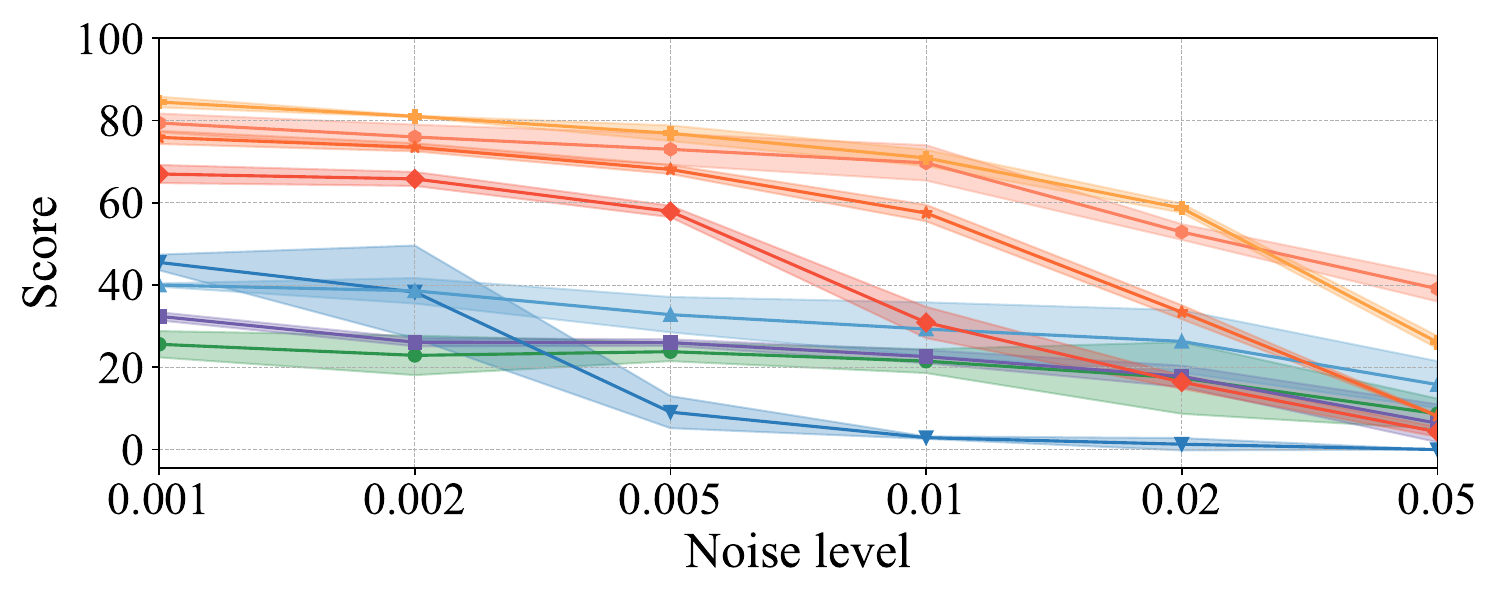}
        \caption{\; ROUGE-L}
    \end{subfigure}
    \begin{subfigure}[b]{1\textwidth}
    \centering
        \includegraphics[width=0.8\textwidth]{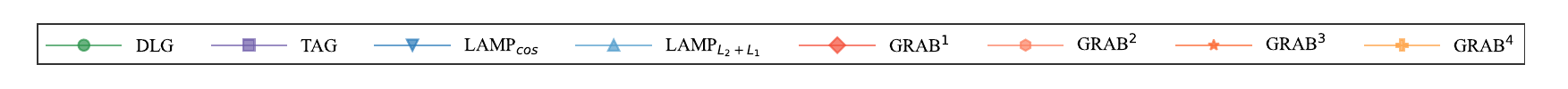}
    \end{subfigure}
    \caption{Performance in practical settings against defense by different gradient noise levels. $\text{\codename}^{1}$: Without noise norm or dropout mask learning. $\text{\codename}^{2}$: Without dropout mask learning. $\text{\codename}^{3}$: Without noise norm. $\text{\codename}^{4}$: With noise norm and dropout mask learning.}
    \label{fig:noise_performance}
\end{figure*}

%% file: tables/noise-tradeoff.tex
\begin{table}[!t]
\centering
\renewcommand{\arraystretch}{0.9}

\caption{DP-SGD utility and privacy trade-off.}
\label{table:noise-tradeoff}
\scriptsize
\resizebox{1\linewidth}{!}{

\setlength{\tabcolsep}{3pt}
\begin{threeparttable}
\begin{tabular}{lrrrrrr}
\toprule
\makecell[c]{Noise level} & \makecell[c]{0.001} & \makecell[c]{0.002} & \makecell[c]{0.005} & \makecell[c]{0.01} & \makecell[c]{0.02} & \makecell[c]{0.05} \\
\midrule
\makecell[c]{MCC} & \makecell[c]{0.74} & \makecell[c]{0.68} & \makecell[c]{0.60} & \makecell[c]{0.51} & \makecell[c]{0.45} & \makecell[c]{0.25} \\
\makecell[c]{$\epsilon$} & \makecell[c]{$1.10 \times 10^{10}$} & \makecell[c]{$2.75 \times 10^{9}$} & \makecell[c]{$4.38 \times 10^{8}$} & \makecell[c]{$1.08 \times 10^{8}$} & \makecell[c]{$2.55 \times 10^{7}$} & \makecell[c]{$2.37 \times 10^{6}$} \\
\bottomrule
\end{tabular}
\end{threeparttable}
}
\vspace{-13pt}
\end{table}

%% file: tables/RQ4_2.tex
\begin{table*}[!t]
\centering
\newcommand{\threecol}[1]{\multicolumn{3}{c}{#1}} 
\newcommand{\skiplen}{0.006\linewidth} 
\renewcommand{\arraystretch}{1.15}

\caption{Performance in practical settings against defense by different gradient pruning ratios.}
\label{table:RQ4_2}

\setlength{\tabcolsep}{3pt}
\begin{threeparttable}
\scriptsize

\begin{tabular}{l rrr p{\skiplen}  rrr p{\skiplen} rrr p{\skiplen}  rrr p{\skiplen}  rrr}
\toprule
& \threecol{Gradient Pruning Ratio: 0.75 (MCC=0.72)} && 
\threecol{Gradient Pruning Ratio: 0.90 (MCC=0.67)} &&
\threecol{Gradient Pruning Ratio: 0.95 (MCC=0.61)} && 
\threecol{Gradient Pruning Ratio: 0.99 (MCC=0.45)} \\
\cmidrule(l{5pt}r{5pt}){2-4} 
\cmidrule(l{5pt}r{5pt}){6-8}
\cmidrule(l{5pt}r{5pt}){10-12}
\cmidrule(l{5pt}r{5pt}){14-16}
& \makecell[c]{R-1} & \makecell[c]{R-2} & \makecell[c]{R-L} && 
\makecell[c]{R-1} & \makecell[c]{R-2} & \makecell[c]{R-L} &&
\makecell[c]{R-1} & \makecell[c]{R-2} & \makecell[c]{R-L} &&
\makecell[c]{R-1} & \makecell[c]{R-2} & \makecell[c]{R-L} \\
\midrule
DLG & 
\makecell[c]{1.1~$\pm$~0.4} & \makecell[c]{0.0~$\pm$~0.0} & \makecell[c]{1.1~$\pm$~0.3} && 
\makecell[c]{0.4~$\pm$~0.2} & \makecell[c]{0.0~$\pm$~0.0} & \makecell[c]{0.4~$\pm$~0.2} &&
\makecell[c]{0.5~$\pm$~0.2} & \makecell[c]{0.0~$\pm$~0.0} & \makecell[c]{0.5~$\pm$~0.2} && 
\makecell[c]{0.4~$\pm$~0.3} & \makecell[c]{0.0~$\pm$~0.0} & \makecell[c]{0.4~$\pm$~0.3} \\

TAG & 
\makecell[c]{0.2~$\pm$~0.3} & \makecell[c]{0.0~$\pm$~0.0} & \makecell[c]{0.2~$\pm$~0.3} && 
\makecell[c]{0.1~$\pm$~0.1} & \makecell[c]{0.0~$\pm$~0.0} & \makecell[c]{0.1~$\pm$~0.1} &&
\makecell[c]{0.2~$\pm$~0.1} & \makecell[c]{0.0~$\pm$~0.0} & \makecell[c]{0.2~$\pm$~0.1} && 
\makecell[c]{0.0~$\pm$~0.1} & \makecell[c]{0.0~$\pm$~0.0} & \makecell[c]{0.0~$\pm$~0.1} \\

LAMP$_{cos}$ & 
\makecell[c]{27.9~$\pm$~6.4\;\;} & \makecell[c]{8.8~$\pm$~2.8} & \makecell[c]{21.3~$\pm$~5.3\;\;} && 
\makecell[c]{7.9~$\pm$~0.8} & \makecell[c]{1.8~$\pm$~0.6} & \makecell[c]{6.7~$\pm$~0.7} &&
\makecell[c]{0.1~$\pm$~0.1} & \makecell[c]{0.0~$\pm$~0.0} & \makecell[c]{0.1~$\pm$~0.1} && 
\makecell[c]{0.0~$\pm$~0.0} & \makecell[c]{0.0~$\pm$~0.0} & \makecell[c]{0.0~$\pm$~0.0} \\

LAMP$_{L_{2}+L_{1}}$ & 
\makecell[c]{0.2~$\pm$~0.2} & \makecell[c]{0.0~$\pm$~0.0} & \makecell[c]{0.2~$\pm$~0.2} && 
\makecell[c]{0.4~$\pm$~0.3} & \makecell[c]{0.0~$\pm$~0.0} & \makecell[c]{0.4~$\pm$~0.3} &&
\makecell[c]{0.3~$\pm$~0.2} & \makecell[c]{0.0~$\pm$~0.0} & \makecell[c]{0.3~$\pm$~0.2} && 
\makecell[c]{0.3~$\pm$~0.1} & \makecell[c]{0.0~$\pm$~0.0} & \makecell[c]{0.3~$\pm$~0.1} \\

\codename$^1$ & 
\makecell[c]{0.9~$\pm$~0.4} & \makecell[c]{0.0~$\pm$~0.0} & \makecell[c]{0.9~$\pm$~0.4} && 
\makecell[c]{0.4~$\pm$~0.2} & \makecell[c]{0.0~$\pm$~0.0} & \makecell[c]{0.4~$\pm$~0.2} &&
\makecell[c]{0.5~$\pm$~0.4} & \makecell[c]{0.0~$\pm$~0.0} & \makecell[c]{0.5~$\pm$~0.4} && 
\makecell[c]{0.3~$\pm$~0.1} & \makecell[c]{0.0~$\pm$~0.0} & \makecell[c]{0.3~$\pm$~0.1} \\

\codename$^2$ & 
\makecell[c]{79.7~$\pm$~1.0\;\;} & \makecell[c]{50.3~$\pm$~1.5\;\;} & \makecell[c]{69.1~$\pm$~1.8\;\;} && 
\makecell[c]{72.7~$\pm$~8.5\;\;} & \makecell[c]{41.2~$\pm$~9.3\;\;} & \makecell[c]{61.2~$\pm$~8.1\;\;} &&
\makecell[c]{76.0~$\pm$~3.6\;\;} & \makecell[c]{45.5~$\pm$~5.2\;\;} & \makecell[c]{65.2~$\pm$~3.7\;\;} && 
\makecell[c]{67.1~$\pm$~3.2\;\;} & \makecell[c]{35.2~$\pm$~4.7\;\;} & \makecell[c]{56.6~$\pm$~2.9\;\;} \\

\codename$^3$ & 
\makecell[c]{0.8~$\pm$~0.1} & \makecell[c]{0.0~$\pm$~0.0} & \makecell[c]{0.7~$\pm$~0.0} && 
\makecell[c]{0.9~$\pm$~0.6} & \makecell[c]{0.0~$\pm$~0.0} & \makecell[c]{0.9~$\pm$~0.6} &&
\makecell[c]{0.9~$\pm$~0.2} & \makecell[c]{0.0~$\pm$~0.1} & \makecell[c]{0.9~$\pm$~0.2} && 
\makecell[c]{0.7~$\pm$~0.3} & \makecell[c]{0.0~$\pm$~0.0} & \makecell[c]{0.7~$\pm$~0.3} \\

\codename$^4$ & 
\makecell[c]{\textbf{89.2}~$\pm$~1.6\;\;} & \makecell[c]{\textbf{65.8}~$\pm$~2.1\;\;} & \makecell[c]{\textbf{78.8}~$\pm$~1.4\;\;} && 
\makecell[c]{\textbf{89.4}~$\pm$~1.6\;\;} & \makecell[c]{\textbf{67.8}~$\pm$~4.4\;\;} & \makecell[c]{\textbf{80.8}~$\pm$~1.9\;\;} &&
\makecell[c]{\textbf{87.6}~$\pm$~0.9\;\;} & \makecell[c]{\textbf{65.3}~$\pm$~2.9\;\;} & \makecell[c]{\textbf{78.0}~$\pm$~2.0\;\;} && 
\makecell[c]{\textbf{85.9}~$\pm$~1.1\;\;} & \makecell[c]{\textbf{60.2}~$\pm$~1.5\;\;} & \makecell[c]{\textbf{75.4}~$\pm$~1.5\;\;} \\

\bottomrule
\end{tabular}
\begin{tablenotes}
\scriptsize
    \item
    $^1$
    Without pruning mask or dropout mask learning.
    $^2$
    Without dropout mask learning.
    $^3$
    Without pruning mask.
    $^4$
    With both pruning mask and dropout mask learning.
\end{tablenotes}
\end{threeparttable}
\vspace{-6pt}
\end{table*}

%% file: sections/6_conclusion_cr.tex
\section{Conclusion}

In this work, we propose \codename, a domain-specific gradient inversion attack with hybrid optimization to recover training data of language models in practical FL scenarios. \codename exploits the gradient decent process during training by incorporating dropout-aware continuous optimization for token recovery, and the beam search discrete optimization for token reordering. 
Our approach effectively recovers a large portion of training data and exhibits notable adaptability to a variety of challenging settings. 
It also demonstrates significant resilience against existing defense mechanisms.

\codename represents a step forward in understanding the privacy threats posed by gradient inversion attacks against the emerging FL training and fine-tuning modes of language models.
However, it currently faces limitations in validating whether the recovered data is part of the training dataset. 
To further enhance \codename, applying techniques such as membership inference~\cite{shokri2017membership, yeom2018privacy, carlini2022membership, mireshghallah2022quantifying, hu2022m} in the context of language models could be explored. 
For example, the loss value of the target model on the recovered data can be utilized to determine the membership threshold~\cite{yeom2018privacy, carlini2021extracting}.

Our findings also highlight the urgent need for the research community to develop robust defense mechanisms to mitigate such threats with consideration of practical training settings and minimum adversary knowledge.
Existing defenses that perturb gradients may mitigate the effectiveness of \codename, as it relies on minimizing the distance between the gradients generated by the original data and those from the recovered data. 
However, these defenses often cause significant model utility loss due to the small magnitudes of these gradients. To successfully defend \codename and similar gradient-based attacks, the defense needs to explore other avenues than perturbing gradients. We suggest potential defenses in two phases of the training, \textit{i.e.,} at the start of the training and during the training process.
\begin{itemize}
    \item At the start of the training, one potential defense is to transform input token embeddings or to substitute the input tokens with different tokens with similar embeddings. 
By doing so, the generated gradients may not be pointed directly to the original tokens.

    \item During the training process, the defense might consider only training and perturbing the gradients of the added layers (\textit{e.g.,} the classification layer) with gradient noise or gradient clipping. Since these layers have gradients with larger magnitudes, higher noise levels may be tolerated.
\end{itemize}

%% file: sections/appendix_A.tex
\section{Supplementary Tables}
\label{appendix_A}

\subsection{FL Setup}
\label{FL_setup}
Table~\ref{table:multi} shows the result of GRAB in benchmark settings on the CoLA dataset with batch sizes from 1 to 32 on $\text{BERT}_{base}$ with multiple (5) clients or a single client. In the case of multiple clients, the victim client possesses the same 64 samples as in the main experiments, and all other clients possess 64 samples that do not overlap. 
The performance remains consistent between the two FL setups, where the attack efficacy of the multiple clients setup is slightly better in most cases. Due to this consistency, we adopt the single client FL setup for simplicity as they are functionally equivalent. 

\subsection{Multi-class Classification}
Table~\ref{table:multi-class} shows the results of \codename in practical settings on the Yahoo Answers Topics dataset with batch sizes from 1 to 32 on $\text{BERT}_{base}$ without the known label assumption. The dataset consists of samples from 10 classes. The evaluation shows that \codename is robust in handling such challenging tasks and can accurately recover both the ground truth labels and the textual data.

\subsection{Large Batch Size}
Table~\ref{table:large-b-size} shows the results of \codename in practical settings on the CoLA dataset with batch sizes from 64 to 128 on $\text{BERT}_{base}$. The evaluation reveals that the attack efficacy of \codename diminishes as the batch size increases. Nevertheless, a substantial amount of private data is still recovered, highlighting \codename's strong recovery capability and underscoring the serious privacy risk it poses, even with large batch sizes.

%% file: sections/appendix_B.tex
\section{Additional Experimental Settings}
\label{appendix_B}

\subsection{Empirical Experiment Implementation}

We implement our approach in Python 3.9.4 with PyTorch 2.0.1.
Different platforms are used to conduct our experiments for practical efficiency reasons.
They include three high-performance computer clusters and two workstations. 
The three high-performance computer clusters run on Rocky Linux with NVIDIA H100 GPUs, CentOS with NVIDIA V100 GPUs, and SUSE Linux with NVIDIA P100 GPUs and NVIDIA H100 GPUs, respectively. 
The two workstations run on Ubuntu with NVIDIA RTX A6000 GPUs and Windows 10 with NVIDIA A4000 GPUs, respectively. 
To ensure consistent results, we test all approaches across all platforms using a single sentence from the CoLA dataset. By setting a fixed random seed, we ensure the consistency and the reproducibility of both intermediate and final outcomes.

\subsection{Model Implementation in Practical Settings}

In the majority of experiments conducted within the practical settings, we run two variants of \codename, \textit{i.e.,} with and without dropout mask learning. For all approaches, the language model of the victim client freezes the embedding layers and activates dropout during training. For all baseline approaches and \codename without dropout mask learning, the language model of the attacker deactivates dropout during recovery, as this serves as a better method when the dropout mask realization of the client is unknown~\cite{scheliga2023dropout}. For \codename with dropout mask learning, the attacker initializes a dropout mask and retains it for optimization.

\subsection{ROUGE Score Calculation}

For the detailed calculation of the ROUGE score, we compare a reference sequence with each recovered sequence in the data batch and select the maximum ROUGE score.
This way ensures the similarity of a recovered data batch with a different sequence order from the reference data batch is obtained. 
Note that in LAMP~\cite{lamp}, the recovered data batch is directly compared with the reference data batch with the same default index, which can result in a miscalculation of the metric.

\input{tables/multi}

\input{tables/multi-class}
\input{tables/large-b-size}

%% file: tables/multi.tex
\begin{table}[t]
\centering
\scriptsize
\newcommand{\threecol}[1]{\multicolumn{3}{c}{#1}} 
\newcommand{\skiplen}{0.004\linewidth} 
\renewcommand{\arraystretch}{1.}
\caption{Performance with multiple clients and a single client.}
\label{table:multi}

\begingroup
\setlength{\tabcolsep}{4pt}

\begin{threeparttable}
\begin{tabular}{l rrr p{\skiplen} rrr p{\skiplen} rrr p{\skiplen} rrr}
\toprule
& \threecol{B = 1 } && \threecol{B = 2 } \\
\cmidrule(l{5pt}r{5pt}){2-4} 
\cmidrule(l{5pt}r{5pt}){6-8} 
& \makecell[c]{R-1} & \makecell[c]{R-2} & \makecell[c]{R-L} && 
\makecell[c]{R-1} & \makecell[c]{R-2} & \makecell[c]{R-L} \\
\midrule
Single 
& 86.4~$\pm$~4.7 & 67.6~$\pm$~9.3 & 78.3~$\pm$~6.2 && 
79.8~$\pm$~2.8 & 57.1~$\pm$~3.6 & 70.0~$\pm$~1.6 \\
Multi 
& \textbf{91.8}~$\pm$~1.9 & \textbf{73.7}~$\pm$~6.7 & \textbf{82.5}~$\pm$~4.3 && 
\textbf{84.4}~$\pm$~3.6 & \textbf{60.7}~$\pm$~8.0 & \textbf{74.1}~$\pm$~5.5 \\
\midrule
& \threecol{B = 4 } && \threecol{B = 8 } \\
\midrule
Single 
& 61.9~$\pm$~3.7 & 35.2~$\pm$~4.9 & 51.5~$\pm$~3.1 &&
53.3~$\pm$~3.5 & \textbf{26.4}~$\pm$~4.3 & 44.7~$\pm$~2.6 \\
Multi 
& \textbf{70.5}~$\pm$~2.9 & \textbf{42.8}~$\pm$~5.2 & \textbf{59.2}~$\pm$~3.8 &&
\textbf{55.8}~$\pm$~3.0 & 25.8~$\pm$~2.5 & \textbf{45.3}~$\pm$~1.6 \\
\midrule
& \threecol{B = 16 } && \threecol{B = 32 } \\
\midrule
Single 
& 45.5~$\pm$~2.2 & 18.1~$\pm$~2.3 & 38.1~$\pm$~2.3 && 
31.9~$\pm$~6.2 & 8.4~$\pm$~3.8 & 28.0~$\pm$~4.6 \\
Multi 
& \textbf{50.2}~$\pm$~4.6 & \textbf{21.8}~$\pm$~3.8 & \textbf{41.3}~$\pm$~3.5 && 
\textbf{34.5}~$\pm$~2.7 & \textbf{11.1}~$\pm$~2.1 & \textbf{30.4}~$\pm$~2.2 \\
\bottomrule
\end{tabular}
\end{threeparttable}
\endgroup
\end{table}

%% file: tables/multi-class.tex
\begin{table}[t]
\centering
\scriptsize
\newcommand{\threecol}[1]{\multicolumn{3}{c}{#1}} 
\newcommand{\skiplen}{0.004\linewidth} 
\renewcommand{\arraystretch}{1.}
\caption{Performance with a 10-class classification task.}
\label{table:multi-class}

\begingroup
\setlength{\tabcolsep}{4pt}

\begin{threeparttable}
\begin{tabular}{l rrr p{\skiplen} rrr p{\skiplen} rrr p{\skiplen} rrr}
\toprule
& \threecol{B = 1 } && \threecol{B = 2 } \\
\cmidrule(l{5pt}r{5pt}){2-4} 
\cmidrule(l{5pt}r{5pt}){6-8} 
& \makecell[c]{R-1} & \makecell[c]{R-2} & \makecell[c]{R-L} && 
\makecell[c]{R-1} & \makecell[c]{R-2} & \makecell[c]{R-L} \\
\midrule
\codename 
& 85.3~$\pm$~1.9 & 54.2~$\pm$~4.1 & 73.6~$\pm$~2.3 && 
74.4~$\pm$~4.2 & 42.1~$\pm$~4.1 & 61.4~$\pm$~3.6 \\

\midrule
& \threecol{B = 4 } && \threecol{B = 8 } \\
\midrule
\codename 
& 62.9~$\pm$~3.7 & 31.4~$\pm$~2.0 & 50.7~$\pm$~2.5 &&
56.1~$\pm$~3.7 & \textbf{23.8}~$\pm$~2.0 & 45.2~$\pm$~3.0 \\

\midrule
& \threecol{B = 16 } && \threecol{B = 32 } \\
\midrule
\codename 
& 53.2~$\pm$~4.2 & 20.0~$\pm$~2.3 & 41.8~$\pm$~2.8 && 
51.3~$\pm$~6.5 & 18.9~$\pm$~4.2 & 41.1~$\pm$~5.5 \\

\bottomrule
\end{tabular}
\end{threeparttable}
\endgroup
\end{table}

%% file: tables/large-b-size.tex
\begin{table}[t]
\centering
\scriptsize
\newcommand{\threecol}[1]{\multicolumn{3}{c}{#1}} 
\newcommand{\skiplen}{0.004\linewidth} 
\renewcommand{\arraystretch}{1.}
\caption{Performance with larger batch sizes.}
\label{table:large-b-size}

\begingroup
\setlength{\tabcolsep}{4pt}

\begin{threeparttable}
\begin{tabular}{l rrr p{\skiplen} rrr p{\skiplen} rrr p{\skiplen} rrr}
\toprule
& \threecol{B = 64 } && \threecol{B = 128 } \\
\cmidrule(l{5pt}r{5pt}){2-4} 
\cmidrule(l{5pt}r{5pt}){6-8} 
& \makecell[c]{R-1} & \makecell[c]{R-2} & \makecell[c]{R-L} && 
\makecell[c]{R-1} & \makecell[c]{R-2} & \makecell[c]{R-L} \\
\midrule
\codename 
& 45.2~$\pm$~2.1 & 19.3~$\pm$~3.5 & 38.9~$\pm$~2.3 && 
39.7~$\pm$~0.7 & 13.6~$\pm$~0.5 & 35.3~$\pm$~1.0 \\



\bottomrule
\end{tabular}
\end{threeparttable}

\endgroup
\end{table}